\documentclass[10pt,journal,compsoc]{IEEEtran}
\usepackage{amsmath,amsfonts}
\usepackage{algorithmic}
\usepackage{algorithm}
\usepackage{array}
\usepackage[caption=false,font=normalsize,labelfont=sf,textfont=sf]{subfig}
\usepackage{textcomp}
\usepackage{stfloats}
\usepackage{url}
\usepackage{verbatim}
\usepackage{graphicx}
\usepackage{multirow}
\usepackage{amssymb}
\usepackage{arydshln}
\usepackage{pifont}
\usepackage{cite}
\usepackage[colorlinks=true,linkcolor=blue,citecolor=blue,urlcolor=blue]{hyperref}
\begin{document}

\title{RegTrack: Simplicity Beneath Complexity in Robust Multi-Modal 3D Multi-Object Tracking}

\author{Lipeng Gu, Xuefeng Yan, Song Wang, \textit{Senior Member}, \textit{IEEE}, Mingqiang Wei, \textit{Senior Member}, \textit{IEEE}
\thanks{L. Gu, M. Wei are with the School of Computer Science and Technology, Nanjing University of Aeronautics
and Astronautics, Nanjing, China, and also with Shenzhen Institute of Research, Nanjing University of Aeronautics
and Astronautics, Shenzhen, China  (e-mail: glp1224@163.com; mingqiang.wei@gmail.com).}
\thanks{X. Yan is with the School of Computer Science and Technology, Nanjing University of Aeronautics and Astronautics, Nanjing, China, and also with the Collaborative Innovation Center of Novel Software Technology and Industrialization, Nanjing, China (e-mail: yxf@nuaa.edu.cn).}
\thanks{S. Wang is with the School of Computer Science and Control Engineering, Shenzhen University of Advanced Technology, Shenzhen, China (e-mail: wangsong@suat-sz.edu.cn).}
}

\markboth{Journal of \LaTeX\ Class Files,~Vol.~14, No.~8, August~2021}%
{Shell \MakeLowercase{\textit{et al.}}: A Sample Article Using IEEEtran.cls for IEEE Journals}


\IEEEtitleabstractindextext{
\begin{abstract}

Existing 3D multi-object tracking (MOT) methods often sacrifice efficiency and generalizability for robustness, largely relying on complex association metrics derived from multi-modal architectures and class-specific motion priors.
Challenging the rooted belief that greater complexity necessarily yields greater robustness, we propose a \textbf{r}obust, \textbf{e}fficient, and \textbf{g}eneralizable method for multi-modal 3D MOT, dubbed \textbf{\textit{RegTrack}}.
Inspired by Yang–Mills gauge theory, RegTrack represents 3D objects using point clouds as \textit{matter fields} and interprets inter-frame object motion as \textit{local variations}. Geometric cues are further modeled as \textit{gauge fields} to adaptively compensate for these variations. In addition, a well-pretrained image representation space serves as a globally invariant \textit{physical law} to guide the compensation process. Consequently, 3D objects represented by point clouds, viewed as \textit{observables}, remain invariant across frames.
Specifically, RegTrack is built upon a unified tri-cue encoder (UTEnc), comprising three tightly coupled components:
a local–global point cloud encoder (LG-PEnc), a mixture-of-experts-based geometry encoder (MoE-GEnc), and an image encoder from a well-pretrained visual-language model.
LG-PEnc efficiently encodes the spatial and structural information of point clouds to produce foundational representations for each object, whose pairwise similarities serve as the sole association metric.
MoE-GEnc seamlessly interacts with LG-PEnc to model inter-object geometric relationships across frames, adaptively compensating for inter-frame object motion without relying on any class-specific priors.
The image encoder is kept frozen and is used exclusively during training to provide a well-pretrained representation space. Point cloud representations are aligned to this space to supervise the motion compensation process, encouraging representation invariance across frames for the same object while enhancing discriminability among different objects.
Through this formulation, RegTrack attains robust, efficient, generalizable inference using only point cloud inputs, requiring just 2.6M parameters.
Extensive experiments on KITTI and nuScenes show that RegTrack outperforms its thirty-five competitors.
\end{abstract}

}

\maketitle

\section{Introduction}

The growing prevalence of 3D sensors, such as LiDAR and RGB-D cameras, has driven the widespread adoption of 3D multi-object tracking (MOT) across diverse domains, including autonomous driving, indoor robotics, and unmanned aerial vehicles \cite{chen2024end,leotta2010vehicle,leibe2008coupled,xu2019dac}.
3D MOT exploits 3D spatial information to enhance the understanding of the physical environment. 
Existing 3D MOT methods can be broadly categorized into single-modal and multi-modal types. 
The former relies on a single sensing modality, while the latter integrates information from multiple complementary sensors.
Regardless of the paradigm, the core challenge lies in designing robust association metrics that match detections to trajectories, ultimately yielding robust 3D trajectories.

\begin{figure}[h]
 \centering
  \includegraphics[width=0.48\textwidth]{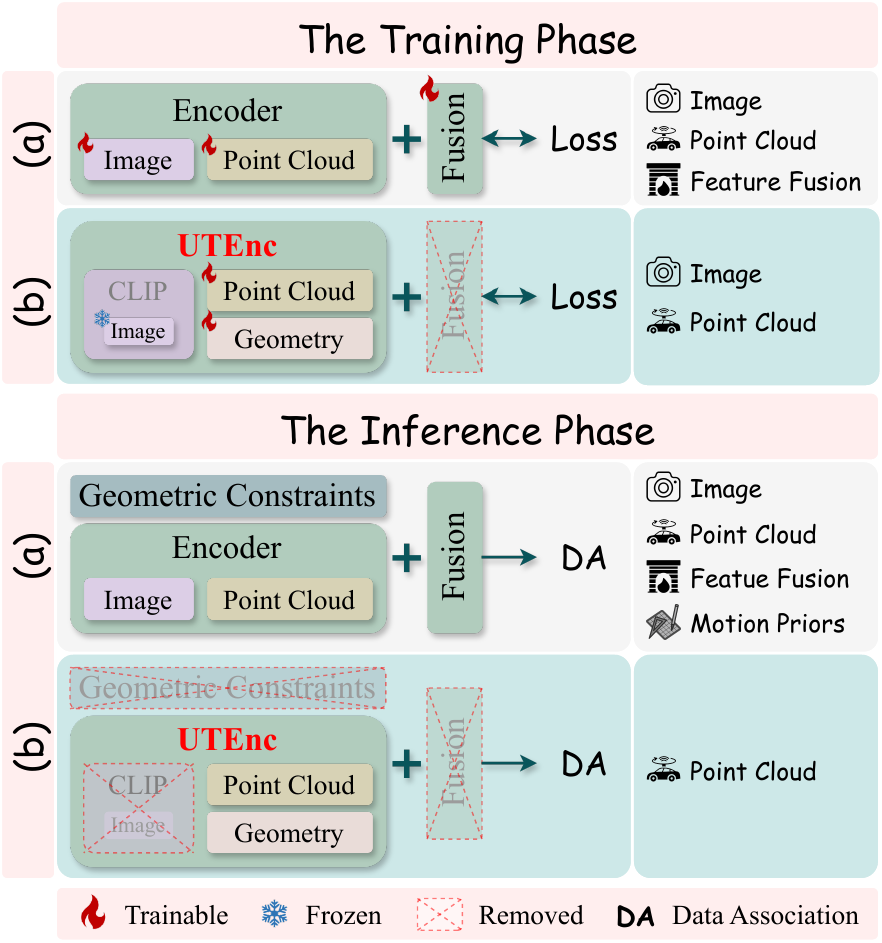}
  \caption{
  \textbf{Comparison between existing methods \cite{jrmot,jmodt,fantrack,mmmot,BcMODT,mmf-jdt} (a) and our RegTrack (b).} 
  i) \textit{Training:}
  RegTrack leverages the representation space of a well-pretrained CLIP image encoder to supervise the joint learning of the point cloud and geometry encoders, thereby facilitating the learning of motion-compensated point cloud representations.
  ii) \textit{Inference:}
  RegTrack employs the point cloud and geometry encoders to encode point cloud inputs, yielding object representations.
  Based on these representations, it constructs a fixed-threshold association metric, achieving superior robustness, efficiency, and generalizability compared with existing methods that rely on intricate multi-modal architectures and geometric constraints driven by class-specific motion priors.
  }
  \label{fig:pipeline}
\end{figure}

\begin{figure*}[th]
 \centering
  \includegraphics[width=\textwidth]{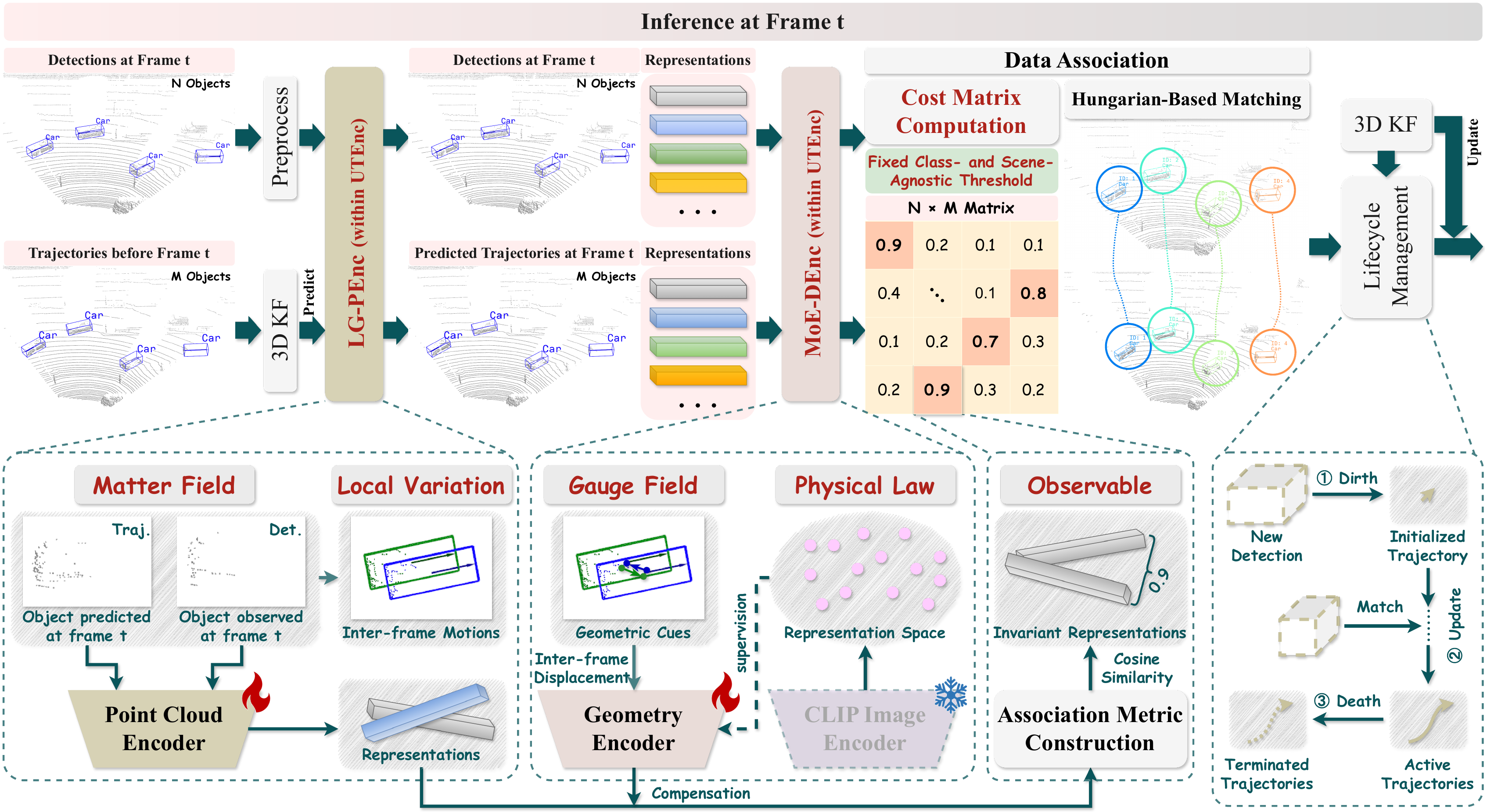}
  \caption{\textbf{Motivation and inference framework of RegTrack.}
  Inspired by Yang–Mills gauge theory, RegTrack employs a geometry encoder to model geometric cues for adaptive motion compensation under the supervision of a CLIP image representation space during training. Consequently, the point cloud encoder learns cross-frame invariant object representations for constructing a robust association metric.
  During inference, RegTrack relies only on the point cloud encoder (LG-PEnc) and the geometric encoder (MoE-GEnc), without requiring the image encoder. Detections are preprocessed by cropping point cloud patches and resampling each patch to $K$ points, while trajectories at frame $t-1$ are propagated to frame $t$ via 3D Kalman filter (KF) prediction (see Fig.~\ref{fig:kf_prediction}). The lifecycle management module handles track birth, update, and death.
  }
  \label{fig:overview}
\end{figure*}

Single-modal methods typically rely on LiDAR point clouds to construct geometric constraints used as association metrics.
These metrics include centroid distance \cite{centerpoint,cbmot,voxelnext}, 3D-IoU \cite{fairmot,ab3dmot}, 3D-GIoU \cite{poly-mot,simpletrack}, A-GIoU \cite{fast-poly}, NN distance \cite{shasta,zaech2022learnable,3dmotformer}, 3D-Box affinity \cite{castrack}.
Notably, these metrics and their corresponding thresholds, which are used to filter out unreliable associations, are typically customized for each object category based on class-specific motion priors. In some cases, they are further adjusted to accommodate particular deployment scenarios. 
However, the inherent complexity of the labor-intensive tuning process limits generalizability, as the metrics and thresholds must be re-tuned whenever a new category or deployment scenario is introduced.
Even with extensive tuning, these single-modal methods still exhibit limited robustness under challenging conditions, such as high-speed motion or dense crowds, often resulting in frequent identity switches and trajectory fragmentation.
To further enhance robustness, multi-modal methods \cite{jrmot,jmodt,fantrack,mmmot,BcMODT,mmf-jdt} (see Fig.~\ref{fig:pipeline}(a)) introduce intricate multi-modal architectures to learn discriminative object features from both images and point clouds. 
The pairwise similarity between these object features can serve as a complementary association metric to geometric constraints.
As a result, association failures caused by fast-moving objects can be effectively alleviated, and erroneous associations between spatially adjacent yet visually dissimilar objects in crowded scenes can be reduced.
However, multi-modal architectures typically comprise independent image and point cloud encoders, together with a multi-modal feature fusion module. This design inevitably introduces additional complexity, which significantly degrades computational efficiency, particularly in crowded scenes with a large number of objects.


\textit{Must robustness inevitably come at the cost of efficiency and generalizability?}
Motivated by PointNN \cite{pointnn}, which achieves remarkable performance in point cloud understanding using only three non-learnable components (FPS, KNN, and pooling), we revisit existing multi-modal 3D MOT frameworks to identify potentially removable redundancies.

We observe that their reliance on class-specific motion priors and complex multi-modal architectures largely arises from motion variations, which differ not only across categories (e.g., moving cars versus walking pedestrians) but also within the same category (e.g., parked cars versus moving cars).
The former necessitates class-specific association metrics and corresponding thresholds, whereas the latter often requires multi-modal cues to handle atypical motion patterns for more reliable associations.
These observations motivate a key hypothesis: if different inter-frame motion patterns are modeled adaptively to compensate for motion variations, class-specific priors become unnecessary, and multi-modal architectures can be substantially simplified.

\textit{Yang–Mills gauge theory inspires the above hypothesis.}
It posits that, regardless of the complexity of \textit{local variations} in \textit{matter fields}, corresponding \textit{gauge fields} can be introduced to compensate for these variations, thereby ensuring that \textit{observables} constructed from the matter and gauge fields remain invariant. Moreover, the \textit{physical laws} governing such compensatory interactions also preserve an invariant form.
As shown in Fig. \ref{fig:overview}, object representations encoded from point clouds can be interpreted as matter fields, while their inter-frame motions correspond to local variations. By constructing appropriate gauge fields and enforcing their governing laws, such variations can be compensated.



We propose a simple yet robust framework for multi-modal 3D MOT, dubbed \textit{RegTrack} (see Fig.~\ref{fig:pipeline}(b) versus (a)).
As illustrated in Fig. \ref{fig:overview}, RegTrack employs a geometry encoder to model inter-frame geometric relationships among objects as gauge fields, and leverages a frozen representation space derived from a well-pretrained CLIP image encoder \cite{clip} as a globally invariant physical law.
Specifically, RegTrack integrates point cloud, geometry, and image encoders into a unified tri-cue encoder (UTEnc), which can be streamlined into a lightweight point-cloud-based inference architecture by discarding the image encoder.
The point cloud encoder is built upon a compact set of MLPs and adaptively aggregates both global and local contextual information to encode the spatial and structural characteristics of point clouds, producing foundational object representations.
To capture object dynamics, the geometry encoder adopts a mixture-of-experts architecture to model inter-frame object displacement, enabling adaptive motion compensation of the point cloud representations.
In addition, the image encoder introduces an auxiliary representation space that serves as a supervisory signal to jointly optimize the point cloud and geometry encoders during training.
As a result, representations corresponding to the same object across frames are encouraged to remain consistent, while the discriminability among different objects is further enhanced.
Consequently, the resulting pairwise similarity between object representations constitutes a robust association metric that generalizes effectively across diverse object categories and scenarios using a fixed class- and scene- agnostic threshold.
Collectively, these innovations establish RegTrack as a pioneering framework that achieves robustness, efficiency, and generalizability in multi-modal 3D MOT.

We comprehensively evaluate RegTrack against thirty-five state-of-the-art competitors on KITTI and nuScenes. Experimental results consistently show that RegTrack achieves superior performance and outperforms all competitors. 
The contributions are summarized as follows:
\begin{itemize}
\item 
We introduce RegTrack, the multi-modal 3D MOT framework inspired by Yang–Mills gauge theory. Built upon UTEnc, it achieves strong robustness without sacrificing efficiency and generalizability.
\item 
UTEnc integrates point cloud, geometric, and image cues to learn discriminative object representations for forming an association metric. Only point clouds are required at inference, yielding improved efficiency.
\item
RegTrack uses a fixed association threshold, independent of class-specific priors, for strong generalizability across diverse categories and scenarios.
\end{itemize}

\section{Related Works}

\subsection{3D Multi-object Tracking}
With the rapid advancement of 3D perception technologies \cite{epnet,epnet++,virconv,largekernel3d,focalformer3d,ctrl,wu2025unsupervised,hisc4d}, existing methods for 3D multi-object tracking (3D MOT) have achieved substantial progress.
According to the number of input modalities, including LiDAR point clouds and camera images, 3D MOT methods can be broadly categorized into single-modal and multi-modal types.


\subsubsection{Single-modal 3D MOT}
Single-modal methods \cite{centerpoint,cbmot,voxelnext,fairmot,ab3dmot,poly-mot,simpletrack,fast-poly,shasta,zaech2022learnable,3dmotformer,castrack} typically construct geometric constraints, such as centroid distance, 3D-IoU, 3D-GIoU, A-GIoU, NN distance, and 3D-Box affinity, as association metrics.
Although simple and computationally efficient, these methods often exhibit limited robustness in crowded scenes or under high-speed motion, leading to frequent identity switches and fragmented trajectories.
Moreover, objects from different categories or deployment scenarios exhibit distinct inter-frame motion patterns. As a result, these methods rely on class-specific motion priors to design case-by-case geometric constraints with manually tuned thresholds, thereby compromising overall generalizability.

\subsubsection{Multi-modal 3D MOT}
To enhance robustness, a variety of multi-modal methods have been proposed, including JRMOT~\cite{jrmot}, JMODT~\cite{jmodt}, FANTrack~\cite{fantrack}, mmMOT~\cite{mmmot}, BcMODT~\cite{BcMODT}, and MMF-JDT~\cite{mmf-jdt}. These methods exploit the complementary information from images and point clouds to introduce additional association cues, thereby alleviating erroneous associations between spatially proximate but visually dissimilar objects, as well as association failures caused by rapid motion. 
However, extracting multi-modal features requires complex architectures involving data preprocessing, separate feature encoding, and multi-modal feature fusion, which substantially limits efficiency. 
As the number of objects in a scene increases, the computational burden grows sharply, further degrading efficiency. Consequently, recent studies \cite{poly-mot,simpletrack,fast-poly} have shown reduced interest in multi-modal methods and instead focus on developing increasingly sophisticated geometric constraints within a single-modality paradigm to enhance tracking robustness.

Our RegTrack shifts 3D MOT research back toward a multi-modal paradigm. Inspired by Yang–Mills gauge theory, it leverages image data only during training as auxiliary supervisory signals, while being simplified into a lightweight point-cloud-based architecture at inference without relying on any class-specific priors. As a result, RegTrack achieves efficient and highly generalizable inference while maintaining strong robustness.



\subsection{Multi-modal Representation Learning}


Given the strong zero-shot generalization capabilities of pretrained vision–language models (VLMs), recent works have transferred their rich image–text knowledge from 2D domains to 3D tasks centered on point clouds.
Early works such as PointCLIP~\cite{pointclip} project point clouds into the CLIP representation space through multi-view rendering, enabling zero-shot 3D recognition and segmentation. CLIP2Point~\cite{clip2point} and CLIP2Scene~\cite{clip2scene} further leverage image–depth joint pretraining and semantically driven alignment to improve point cloud classification and scene understanding. These advances have also facilitated new tasks, including 3D anomaly detection~\cite{pointad}, 3D vision–language grounding~\cite{cross3dvg}, open-world 3D understanding~\cite{duoduo-clip,clip-gs,triclip-3d}, and open-vocabulary 3D object detection \cite{findnpropagate,CoDAv2}.

More recently, pretrained VLMs have been introduced into the 3D MOT domain.
Open3DTrack~\cite{open3dtrack} is the first to extend 3D MOT to open-world scenarios and demonstrates strong generalization to unseen categories. However, such recognition capability can already be effectively achieved by open-vocabulary 3D object detection methods. In contrast, Open3DTrack does not adequately take into account two fundamental requirements of 3D MOT: computational efficiency and the generalizability of association metrics and their corresponding thresholds.

Our RegTrack introduces a new perspective that leverages pretrained VLMs to address the long-standing trade-off among robustness, efficiency, and generalizability.

\begin{figure*}
    \centering
    \includegraphics[width=\linewidth]{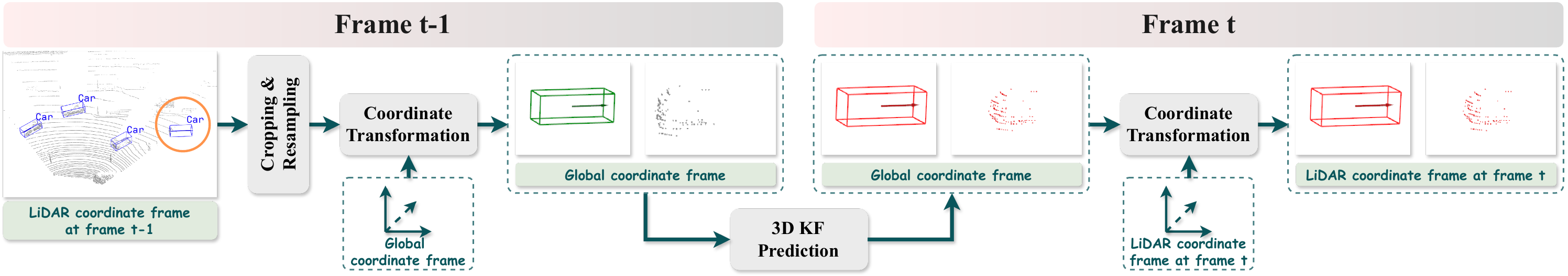}
    \caption{
    \textbf{Pipeline of 3D Kalman filter prediction.}
    Taking an object from a trajectory at frame $t-1$ as an example, the object is cropped into a point cloud patch and resampled to $K$ points. 
    The patch and its 3D bounding box are transformed into the global coordinate frame and propagated to frame $t$ via a 3D Kalman filter prediction. The predicted patch and bounding box are then transformed into the LiDAR coordinate frame at frame $t$.
    }
    \label{fig:kf_prediction}
\end{figure*}


\begin{figure*}[ht]
    \centering
    \includegraphics[width=\linewidth]{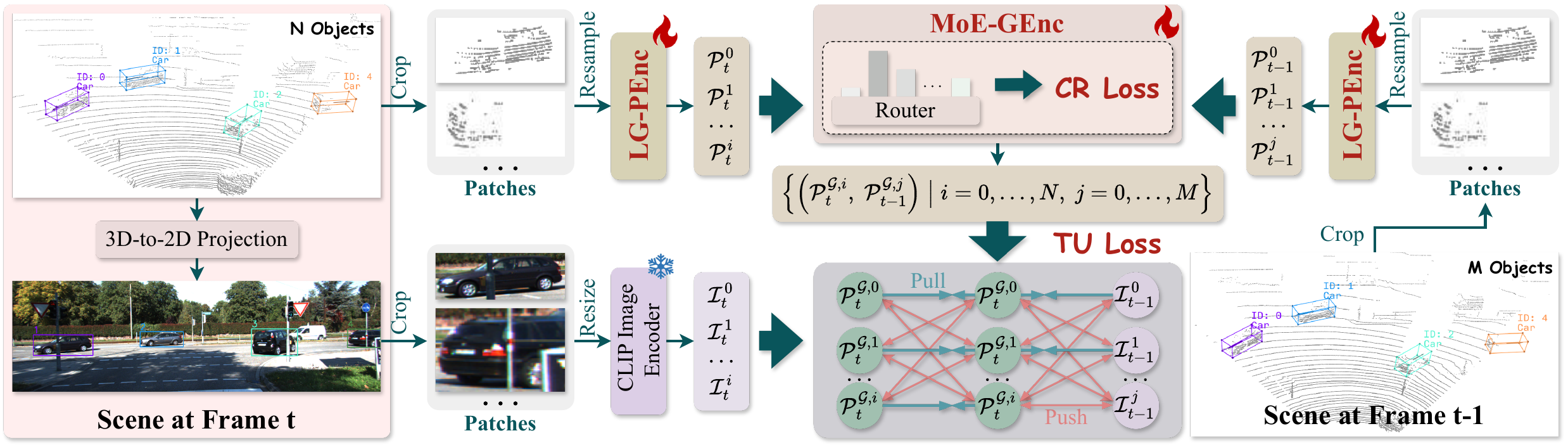}
    \caption{\textbf{Training pipeline of UTEnc.} 
    UTEnc consists of three sub-modules: a local–global point cloud encoder (LG-PEnc), a mixture-of-experts-based geometry encoder (MoE-GEnc), and a well-pretrained CLIP image encoder.
    LG-PEnc encodes point cloud objects to produce object representations $\mathcal{P}$.
    MoE-GEnc then performs adaptive motion compensation on a pair of inter-frame point cloud representations $\mathcal{P}_{t-1}$ and $\mathcal{P}_{t}$, guided by the composite routing (CR) loss, yielding motion-compensated representations $\mathcal{P}^{\mathcal{G}}_{t-1}$ and $\mathcal{P}^{\mathcal{G}}_{t}$.
    The image encoder is frozen and used only during training to provide a globally invariant reference space that supervises the compensation process via the tri-cue unification (TU) loss.
    }
    \label{fig:UTEnc}
\end{figure*}

\section{Method}

\subsection{Overview}
Existing multi-modal 3D MOT methods~\cite{mmmot,jmodt,jrmot,fantrack,BcMODT,mmf-jdt} often employ intricate multi-modal architectures together with class-specific motion priors to construct complex association metrics for robust performance.
However, the substantial computational overhead introduced by multi-modal architectures limits efficiency, while the reliance on class-specific priors constrains generalizability.
These limitations motivate us to revisit existing multi-modal frameworks to identify potentially removable redundancies, achieving robust performance without sacrificing efficiency and generalizability.

Inspired by Yang–Mills gauge theory, we propose a novel multi-modal 3D MOT framework, dubbed RegTrack (see Fig. \ref{fig:pipeline}(b) and Fig. \ref{fig:overview}), which effectively addresses the long-standing trade-off among robustness, efficiency, and generalizability.
RegTrack features a concise architecture built upon a unified tri-cue encoder (UTEnc, see Fig. \ref{fig:UTEnc}).
UTEnc learns discriminative object representations to construct a robust, class-agnostic association metric, enabling reliable inter-frame association between trajectories and detections.
Furthermore, standard data association (DA), 3D Kalman filter (KF), and lifecycle management modules are integrated to complete the tracking pipeline.

As illustrated in Fig. \ref{fig:overview}, the 3D KF consists of prediction and update steps, which propagate trajectories from the previous frame to the current frame and update trajectories using the matched detections, respectively. The lifecycle management module is responsible for track birth, update, and death.
Both modules follow AB3DMOT~\cite{ab3dmot}, and their implementation details are omitted for brevity.

\subsection{Problem Formulation} \label{section: problem formulation}
As illustrated in Fig. \ref{fig:overview}, at frame $t$, the inputs consist of 3D detections $\mathcal{D}_{t}$ and 3D trajectories $\mathcal{T}_{t-1}$, and the output is the updated 3D trajectories $\mathcal{T}_{t}$.
First, the 3D KF propagates $\mathcal{T}_{t-1}$, including the 3D bounding boxes and their corresponding point cloud patches, to frame $t$, yielding the predicted 3D trajectories $\mathcal{\hat{T}}_{t}$ (see Fig. \ref{fig:kf_prediction}).
Next, UTEnc extracts object representations from both $\mathcal{D}_{t}$ and $\hat{\mathcal{T}}_{t}$. 
Cosine similarities between the object representations of 3D detections and predicted 3D trajectories are then computed to construct the cost matrix $\mathcal{C}$.
Based on this cost matrix, the data association module matches 3D detections to predicted 3D trajectories across frames, yielding the updated 3D trajectories $\mathcal{T}_{t}$ and updating the 3D KF accordingly.

\subsection{Unified Tri-cue Encoder} \label{section: Unified Tri-modal Encoder}
Fig.~\ref{fig:pipeline}(a) shows that existing multi-modal 3D MOT methods \cite{jrmot,jmodt,fantrack,mmmot,BcMODT,mmf-jdt} typically rely on intricate multi-modal architectures, which incur substantial computational overhead and significantly reduce efficiency.
Moreover, geometric constraints inherited from single-modal methods are often tied to class-specific priors, necessitating customized designs and threshold tuning for different object categories and even across distinct scenarios, thereby limiting generalizability.

To overcome these limitations, we introduce a unified tri-cue encoder (UTEnc) inspired by Yang–Mills gauge theory. As illustrated in Fig. \ref{fig:overview}, UTEnc comprises three encoders: a local–global point cloud encoder (LG-PEnc), a mixture-of-experts-based geometry encoder (MoE-GEnc), and a well-pretrained CLIP image encoder. 
Without relying on any class-specific prior knowledge, UTEnc leverages geometric cues to adaptively handle diverse inter-frame motion patterns, thereby producing motion-compensated point cloud representations that can be regarded as cross-frame invariant observations. Consequently, the pairwise cosine similarity between these representations serves as a robust, class-agnostic association metric, ensuring the robustness and generalizability of 3D MOT. 
More importantly, the image encoder in UTEnc is used only during training to provide a globally invariant constraint for motion compensation, effectively serving as a physical law, and is discarded during inference, which further improves the efficiency of 3D MOT.


\begin{figure}[t]
    \centering
    \includegraphics[width=1\linewidth]{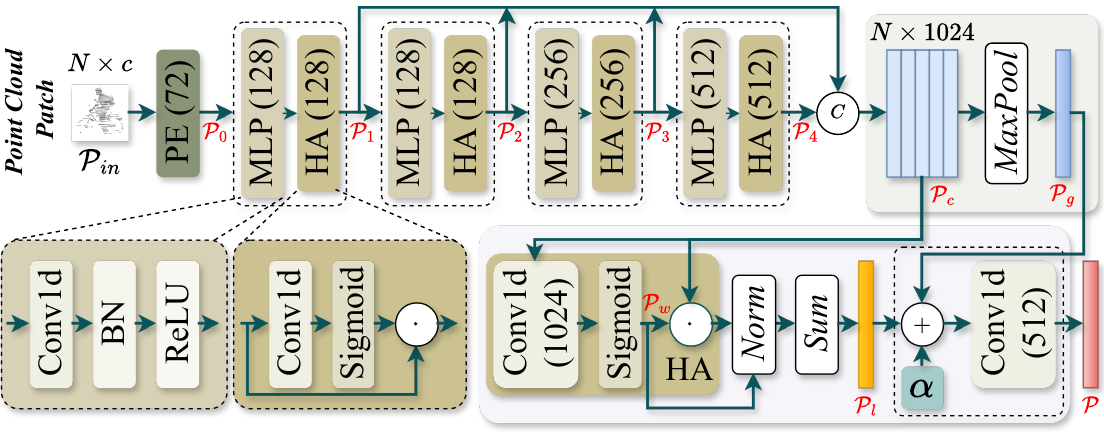}
    \caption{
    \textbf{Overview of LG-PEnc.} 
    It takes a point cloud patch $\mathcal{P}_{in} \in \mathbb R^{N \times c}$ and extracts global $\mathcal{P}_g \in \mathbb R^{1 \times 1024}$ and local $\mathcal{P}_l \in \mathbb R^{1 \times 1024}$ features. 
    The two features are aggregated using a learnable parameter $\alpha \in \mathbb R^{1 \times 1024}$ and a 1D convolution to output the object representation $\mathcal{P} \in \mathbb R^{1 \times 512}$.
    }
    \label{fig:LGPEnc}
\end{figure}

\begin{figure}[t]
    \centering
    \includegraphics[width=1\linewidth]{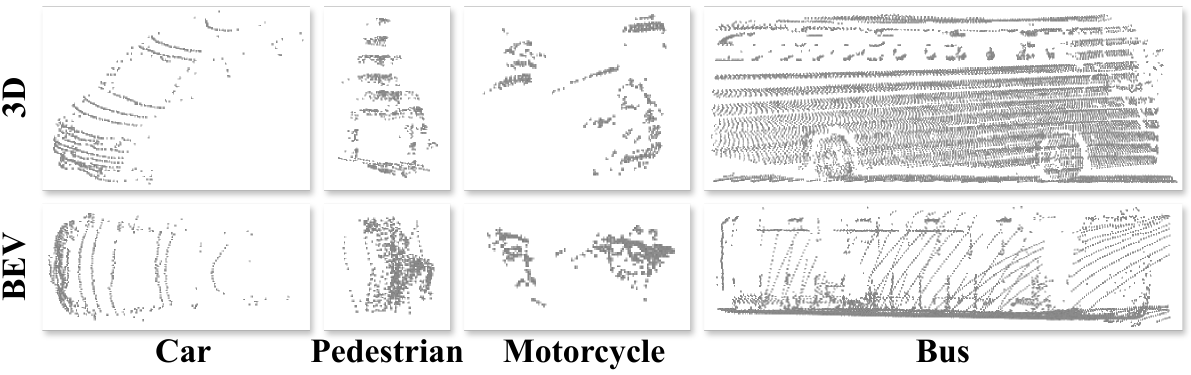}
    \caption{
    \textbf{Examples of objects in 3D space and BEV.} 
    Even when using only the \textit{x} and \textit{y} coordinates in BEV, the contour and appearance of different object categories remain clearly distinguishable, providing sufficient information to support 3D MOT while avoiding potential interference from redundant height information (\textit{z} coordinate).
    }
    \label{fig:object_samples}
\end{figure}

\subsubsection{LG-PEnc} \label{section: LGPEnc}
Advanced point cloud encoders \cite{pointnet,pointnet++,pointconv,dgcnn,pointmixer} face excessive computational overhead when deployed in complex scenes containing large numbers of objects, such as those in the nuScenes autonomous driving dataset \cite{nuscenes}.
To address this limitation, we introduce LG-PEnc (see Fig. \ref{fig:LGPEnc}), which captures complex point cloud characteristics through a lightweight architecture that integrates local and global patterns.
It consists of four MLP layers, five hybrid attention (HA) layers, one learnable parameter, one 1D convolution, and several non-learnable components (e.g., positional encodings and sigmoid functions). Each HA layer employs a 1D convolution together with a sigmoid function to jointly model inter-channel and inter-point dependencies, thereby capturing fine-grained details.
Specifically, a resampled point cloud patch $\mathcal{P}_{in} \in \mathbb{R}^{N \times c}$, containing $N$ points and $c$ channels, is first processed by a positional encoding (PE) module to produce the feature $\mathcal{P}_0 \in \mathbb{R}^{1 \times 72}$. 
The resulting feature is then passed through four MLP layers equipped with HA layers to generate intermediate features $\left \{ \mathcal{P}_1, \mathcal{P}_2, \mathcal{P}_3, \mathcal{P}_4 \right \} $
These features are concatenated along the channel dimension to form the aggregated feature $\mathcal{P}_c \in \mathbb{R}^{N \times 1024}$.
The global feature is then generated as
\begin{equation}
\mathcal{P}_{g} = \mathrm{MaxPool}(\mathcal{P}_{c}) \in \mathbb{R}^{1 \times 1024}.
\end{equation}
The local feature $\mathcal{P}_{l}$ is obtained via an HA layer, defined as
\begin{equation} \label{eq:local features}
\mathcal{P}_{l} = \sum_{i=1}^{N} \frac{\mathcal{P}_{c}(i,j) \ast \mathcal{P}_{w}(i,j)}{\sum_{j=1}^{1024} \mathcal{P}_{w}(i,j)} \in \mathbb{R}^{1 \times 1024}, 
\end{equation}
where the weights $\mathcal{P}_{w} \in \mathbb{R}^{N \times 1024}$ are learned from the features $\mathcal{P}_c$ through the HA layer and are used to adaptively characterize the relative importance of different semantic channels at each point.
Finally, the global and local features are aggregated via an adaptive weighted summation with a learnable parameter $\alpha \in \mathbb{R}^{1 \times 1024}$ and a 1D convolution, yielding the point cloud representation, formulated as
\begin{equation} \label{eq:point cloud embedding}
\mathcal{P} = \mathrm{Conv1d}\left(\alpha \cdot \mathcal{P}_{g} + (1 - \alpha) \cdot \mathcal{P}_{l}\right) \in \mathbb{R}^{1 \times 512}.
\end{equation}

\textbf{Remark 1.} 
For an input point cloud patch $\mathcal{P}_{in} \in \mathbb{R}^{N \times c}$, the second dimension retains only the $x$ and $y$ coordinates in BEV. 
As illustrated in Fig. \ref{fig:object_samples}, this choice is motivated by the observation that tracked objects in autonomous driving scenarios are located on the ground plane. Consequently, height information is not essential for accurately preserving object contours and spatial localization. Moreover, removing the height channel helps suppress potential redundancy, leading to more discriminative object representations. 
These observations are validated experimentally (see Table~\ref{tab4}).

\textbf{Remark 2.} 
The PE module adopts a non-learnable sinusoidal formulation instead of the commonly used learnable MLP-based embedding. This choice is motivated by the observation that sinusoidal positional encoding provides a deterministic mapping that preserves precise spatial location information, generalizes reliably to unseen spatial locations, and mitigates the risk of overfitting typically introduced by learnable embeddings. These observations are further validated by experimental results (see Table~\ref{tab4}).

\textbf{Remark 3.} 
The attention weights for local features are normalized across channels (see Eq.~\ref{eq:local features}), rather than using the commonly adopted point-wise normalization~\cite{fastpillars}.
This choice is motivated by empirical observations in LiDAR point clouds, where 3D objects are represented by a limited number of points that should be treated with equal importance. Consequently, competition among points is considered undesirable. Instead, competition is introduced across feature channels, which facilitates selective semantic activation while avoiding trivial scaling effects.
Moreover, to better aggregate informative cues from both local and global features, adaptive channel-wise weighted summation (see Eq. \ref{eq:point cloud embedding}) is applied instead of fixed average summation.
These observations are validated experimentally (see Table~\ref{tab6-2}).

\begin{figure}[t]
    \centering
    \includegraphics[width=
    \linewidth]{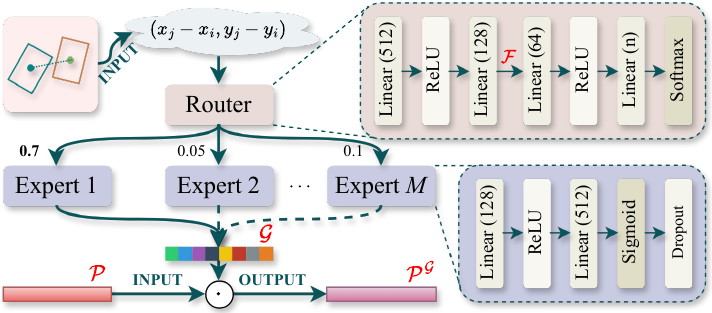}
    \caption{
    \textbf{Overview of MoE-GEnc.}
    It takes as input the relative coordinates $\left ( x_j - x_i, y_j - y_i \right ) $ between a detection $\mathbf{\text{d}}_{i} \in \mathcal{D}_{t}$ and a trajectory $\mathbf{\hat{\text{t}}}_{j} \in \mathcal{\hat{T}}_{t}$ in BEV. 
    A lightweight router then predicts expert-selection probabilities, from which the expert with the highest probability generates a geometric representation $\mathcal{G} \in \mathbb{R}^{1 \times 512}$. 
    Finally, the point cloud representation of detection $\mathbf{d}_i$, denoted as $\mathcal{P} \in \mathbb{R}^{1 \times 512}$, is motion compensated through element-wise multiplication with $\mathcal{G}$, yielding $\mathcal{P}^\mathcal{G} \in \mathbb{R}^{1 \times 512}$.
    }
    \label{fig:MoE-GEnc}
\end{figure}

\subsubsection{MoE-GEnc} \label{section: MoE-GEnc}
Diverse inter-frame motion patterns often reduce the discriminability of the point cloud representations $\mathcal{P}$ generated by LG-PEnc. 
For example, fast-moving objects tend to exhibit low representation similarity across frames due to large spatial displacements and partial appearance changes in LiDAR point clouds, whereas distinct objects that are spatially proximate may undesirably show high inter-frame similarity.
As a result, different motion patterns necessitate different association thresholds, which increase manual tuning effort and degrade model generalizability.
To address this challenge, we draw inspiration from Yang–Mills gauge theory, in which gauge fields compensate for local variations while preserving the invariance of physical observables. Accordingly, we propose MoE-GEnc to adaptively compensate for diverse inter-frame motions, thereby promoting cross-frame invariance of point cloud representations.

As shown in Fig.~\ref{fig:MoE-GEnc}, MoE-GEnc comprises a lightweight router and a set of experts with identical architectures.
Conceptually, it is a form of spatial attention that suppresses the similarity between objects that are visually similar but spatially distant, while enhancing the similarity between objects that are spatially proximate but exhibit limited appearance similarity.
Specifically, MoE-GEnc takes as input a detection–trajectory pair $(\mathbf{d}_i, \hat{\mathbf{t}}_j)$, where $\mathbf{d}_i \in \mathcal{D}_t$ and $\hat{\mathbf{t}}_j \in \hat{\mathcal{T}}_t$.
When performing motion compensation for the detection $\mathbf{d}_{i}$, spatial proximity is first modeled using the relative coordinates $\left(x_{j} - x_{i}, y_{j} - y_{i}\right)$. Conversely, when compensating the trajectory $\hat{\mathbf{t}}{j}$, the relative coordinates $\left(x_{i} - x_{j},, y_{i} - y_{j}\right)$ are used. The router then adaptively selects an appropriate expert to learn inter-frame geometric relationships and encodes them into a geometric representation $\mathcal{G}$, which is applied to motion-compensate the point cloud representation $\mathcal{P}$ of the detection or trajectory, yielding $\mathcal{P}^{\mathcal{G}}$.

Formally, as illustrated in Fig. \ref{fig:MoE-GEnc}, MoE-GEnc operates as
\begin{equation} \label{geometry-enhanced point cloud emebedding}
\begin{aligned}
& \Delta \mathcal{C} = (x_j - x_i, \, y_j- y_i) \in \mathbb{R}^{1 \times 2}, \\
& \mathbf{p} = \mathrm{Router}(\Delta \mathcal{C}) \in \mathbb{R}^{M}, \\
& \mathcal{G} = E_k(\Delta \mathcal{C}) \in \mathbb{R}^{1 \times 512}, \quad \text{where } k = \arg\max_m \mathbf{p}_m, \\
& \mathcal{P^G} = \mathcal{P} \odot \mathcal{G} \in \mathbb{R}^{1 \times 512},
\end{aligned}
\end{equation}
where $\mathrm{Router}(\cdot)$ outputs a probability distribution $\mathbf{p}$ over $M$ experts $\{E_m\}_{m=1}^{M}$;
$\mathcal{G}$ is the geometric representation generated by the selected expert $E_k$ with the highest probability;
and $\odot$ denotes element-wise multiplication.

Finally, the compensated point cloud representation $\mathcal{P^G}$ is passed through a learnable projection layer and subsequently $L2$-normalized, as defined by
\begin{equation} \label{eq:pc representation}
\mathcal{P^G} = norm \left ( \mathcal{P^G} \otimes \mathcal{W}_{\text{proj}} \right ) \in \mathbb{R}^{1 \times 512},
\end{equation} 
where $\mathcal{W}_{\text{proj}} \in \mathbb{R}^{512 \times 512}$ is a projection matrix, and the output is normalized using the $L2$ norm.


\subsubsection{Image Encoder} \label{section: image encoder}
Beyond stating that gauge fields compensate for local variations, Yang–Mills gauge theory further emphasizes that such compensations are governed by underlying physical laws.
Analogously, we adopt the image representation space of the pretrained CLIP model as the governing law that guides the compensation mechanism implemented by MoE-GEnc.
Specifically, the CLIP image encoder extracts image representations $\mathcal{I} \in \mathbb{R}^{1 \times 512}$ from cropped and resized image patches for each object. These image representations are aligned with the corresponding motion-compensated point cloud representation $\mathcal{P}^{\mathcal{G}} \in \mathbb{R}^{1 \times 512}$, which enforces LG-PEnc equipped with MoE-GEnc to be optimized under the supervision of the well-pretrained image representation space.
Note that the image encoder is frozen during training and removed at inference to improve computational efficiency.

\subsection{Data Association} \label{section: Data Association}
Our data association module is deliberately simple.
It uses the similarities between motion-compensated point cloud representations (see Eq. \ref{eq:pc representation}) as the sole association metric. 
Specifically, it constructs the cost matrix $\mathcal{C}$ between 3D detections in $\mathcal{D}_t$ and predicted 3D trajectories in $\hat{\mathcal{T}}_t$ across adjacent frames, as follows:
\begin{equation} \label{cost matrix}
\mathcal{C}=
\begin{bmatrix}
c_{1,1}  & \cdots   & c_{1,N}   \\
c_{2,1}  & \cdots   & c_{2,N}  \\
\vdots   & \ddots   & \vdots  \\
c_{M,1}   & \cdots\  & c_{N,N}  \\
\end{bmatrix},
\end{equation}
\begin{equation}
    c_{i,j} = 1 - \left ( \mathcal{P}^{\mathcal{G},i} \right )^\top \cdot \mathcal{P}^{\mathcal{G},j} \in \left [ -1, 1 \right ] ,
\end{equation}
where $M$ and $N$ denote the number of trajectories in $\mathcal{T}_{t-1}$ and detections in $\mathcal{D}_{t}$, respectively.
Based on this cost matrix, the module then employs the Hungarian algorithm to associate trajectories in $\mathcal{T}_{t-1}$ with detections in $\mathcal{D}_{t}$, producing high-quality 3D trajectories in $\mathcal{T}_{t}$ at frame $t$.

\subsection{Training Loss} \label{section: training loss}

\subsubsection{Composite Routing Loss} \label{section: composite routing loss}
We employ a composite routing (CR) loss to supervise the optimization of MoE-GEnc during training, following prior work \cite{diversity_loss,important_loss,david2014learning}.
This loss jointly enforces expert balance and diversity, thereby promoting both balanced expert utilization and discriminative specialization. It is defined as
\begin{equation}
\mathcal{L}_{\text{cr}} =
\lambda_{\text{bal}}\, \mathcal{L}_{\text{bal}} +
\lambda_{\text{div}}\,
\big(
\mathcal{L}_{\text{intra}} + 
\mathcal{L}_{\text{cent}}
\big),
\label{eq:router_total}
\end{equation}
where $\lambda_{\text{bal}}$ and $\lambda_{\text{div}}$ denote the weighting coefficients for the balance and diversity terms, respectively, and $\mathcal{L}_{\text{bal}}$, $\mathcal{L}_{\text{intra}}$, and $\mathcal{L}_{\text{cent}}$ represent the balance loss, intra-expert compactness loss, and inter-expert diversity loss, respectively.

$\mathcal{L}_{\text{bal}}$ encourages uniform expert utilization across each mini-batch, which mitigates routing collapse and stabilizes the training dynamics of the MoE architecture:
\begin{equation}
\mathcal{L}_{\text{bal}} 
= \frac{1}{E} \sum_{e=1}^{E} 
\left| 
\frac{1}{E} - 
\frac{1}{N_e} \sum_{i=1}^{N_e} p_{i,e} 
\right|,
\label{eq:router_balance}
\end{equation}
where $p_{i,e}$ is the probability of the $i$-th sample assigned to expert $e$, $N_e$ is the number of samples routed to expert $e$, and $E$ is the total number of experts.

$\mathcal{L}_{\text{intra}}$ enforces feature coherence within each expert (i.e., $\mathcal{F}$ in Fig. \ref{fig:MoE-GEnc}) by minimizing the average pairwise dissimilarity among its assigned samples:
\begin{equation}
\mathcal{L}_{\text{intra}} 
= \frac{1}{E} 
\sum_{e=1}^{E} 
\frac{1}{|\mathcal{S}_e|^2}
\sum_{i,j \in \mathcal{S}_e}
\big(1 - 
\cos(\mathcal{F}_{i}, \mathcal{F}_{j})\big),
\label{eq:router_intra}
\end{equation}
where $\mathcal{S}_e$ is the set of samples routed to expert $e$, $|\mathcal{S}_e|^2$ denotes the total number of sample pairs in $\mathcal{S}_e$, and $\cos \left(\cdot, \cdot \right) $ denotes cosine similarity between features.

Finally, $\mathcal{L}_{\text{cent}}$ promotes complementary specializations across experts by minimizing the cosine similarity among their centroids:
\begin{equation}
\mathcal{L}_{\text{cent}} 
= \frac{1}{E(E-1)}
\sum_{\substack{e,e'=1\\e\ne e'}}^{E}
\cos(\mathbf{c}_e, \mathbf{c}_{e'}),
\quad 
\mathbf{c}_e = 
\frac{1}{|\mathcal{S}_e|} 
\sum_{i\in\mathcal{S}_e}\mathcal{F}_i,
\label{eq:router_centroid}
\end{equation}
where $\mathbf{c}_e$ is the centroid feature of expert $e$, and $|\mathcal{S}_e|$ represents the number of samples in $\mathcal{S}_e$.

\begin{figure}
    \centering
    \includegraphics[width=1\linewidth]{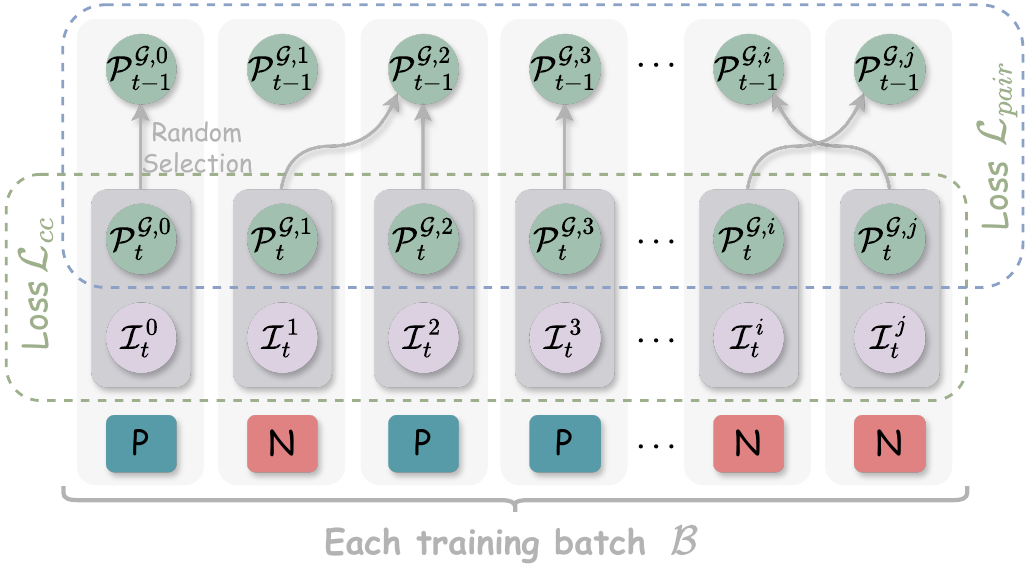}
    \caption{
    \textbf{Illustration of the sampling strategy and the tri-cue unification loss during training.}
    In each training batch, a sample pair $(\mathcal{P}^{\mathcal{G},i}_{t}, \mathcal{I}^{i}_{t}; \mathcal{P}^{\mathcal{G},j}_{t-1})$ is constructed from frames $t$ and $t-1$.
    The pair is positive if $i=j$ and negative otherwise.
    The tri-cue unification loss consists of losses $\mathcal{L}_{cc}$ and $\mathcal{L}_{pair}$ (see Eq.~\ref{eq:tu loss}).
    }
    \label{fig:sampling}
\end{figure}

\subsubsection{Tri-cue Unification Loss} \label{section: UTCL loss}

To better leverage the well-pretrained image representation space to guide MoE-GEnc in performing adaptive inter-frame motion compensation on point cloud representations generated by LG-PEnc, we introduce a tri-cue unification (TU) loss $\mathcal{L}_{tu}$, which integrates a cross-modal contrastive loss $\mathcal{L}_{cc}$ and a pairwise distance loss $\mathcal{L}_{pair}$.
Specifically, $\mathcal{L}_{cc}$ pulls motion-compensated point cloud representations toward their corresponding image representations while pushing them away from non-corresponding ones. Moreover, $\mathcal{L}_{pair}$ further enhances discriminability by encouraging similarity among motion-compensated point cloud representations of the same object and dissimilarity among those of different objects.
Under the joint supervision of $\mathcal{L}_{cc}$ and $\mathcal{L}_{pair}$, LG-PEnc equipped with MoE-GEnc produces more discriminative object representations for robust association metric construction.

To better facilitate training with the TU loss $\mathcal{L}_{tu}$, the sample pairs in each training batch $\mathcal{B}$ are constructed using the sampling strategy illustrated in Fig. \ref{fig:sampling}.
Specifically, in each sample pairs $(\mathcal{P}^{\mathcal{G},i}_{t}, \mathcal{I}^{i}_{t}; \mathcal{P}^{\mathcal{G},j}_{t-1})$, an object sample $(\mathcal{P}^{\mathcal{G},i}_{t}, \mathcal{I}^{i}_{t})$ is randomly sampled from the current frame $t$, and anthor object sample $\mathcal{P}^{\mathcal{G},j}_{t-1}$ is randomly sampled from the previous frame $t-1$. If the two samples correspond to the same object (i.e., $i=j$), the sample pair $(\mathcal{P}^{\mathcal{G},i}_{t}, \mathcal{I}^{i}_{t}; \mathcal{P}^{\mathcal{G},j}_{t-1})$ is treated as a positive pair; otherwise, it is treated as a negative pair.

Formally, the TU loss $\mathcal{L}_{\mathrm{tu}}$ is defined as
\begin{equation} \label{eq:tu loss}
\begin{aligned}
\mathcal{L}_{\mathrm{tu}}
&= 
\gamma \ast \mathcal{L}_{cc}(\mathcal{P}_{t},\mathcal{G}_{t},\mathcal{I}_{t})
+\delta \ast \mathcal{L}_{\mathrm{pair}}\!\big(\mathcal{P}_{t-1},\mathcal{P}_{t},\mathcal{G}_{t-1},\mathcal{G}_{t}\big)  \\
& =
\gamma \ast \mathcal{L}_{cc}(\mathcal{P}^{\mathcal{G}}_{t},\mathcal{I}_{t})
+\delta \ast \mathcal{L}_{\mathrm{pair}}\!\big(\mathcal{P^G}_{t-1},\mathcal{P^{G}}_{t}\big), 
\end{aligned}
\end{equation}
where $\mathcal{P}$, $\mathcal{I}$, and $\mathcal{G}$ denote the point cloud, geometry, and image representations of object samples in a training batch $\mathcal{B}$, respectively. 
$\mathcal{P}^{\mathcal{G}}$ represents the motion-compensated point cloud representation of object samples obtained from Eq. \ref{geometry-enhanced point cloud emebedding}.
The subscripts ``$t-1$'' and ``$t$'' denote that the object samples are drawn from the scenes at frame $t-1$ and frame $t$, respectively, as shown in Fig. \ref{fig:UTEnc}.
The coefficients $\gamma$ and $\delta$ balance the contributions of the losses $\mathcal{L}_{cc}$ and $\mathcal{L}_{pair}$.

The loss $\mathcal{L}_{cc}$ is formulated as
\begin{equation} \label{geometry-enhanced point cloud emebedding}
\begin{aligned}
& \mathcal{L}_{cc}
= -\frac{1}{2}\sum_{i} \Bigg ( \log \frac{\exp\left(s_{i,i}\right)}
{\sum_{j} \exp\left(s_{i,j}\right)}+
\log \frac{\exp\left(s_{i,i}\right)}
{\sum_{j} \exp\left(s_{j,i}\right)}
\Bigg), \\
& s_{i,j} = \frac{(\mathcal{P}^{\mathcal{G},i}_{t})^\top \mathcal{I}^j_{t}}{\tau},
\end{aligned}
\end{equation}
where $i=j$ indicate that $\mathcal{P}^{\mathcal{G},i}_{t}$ and $\mathcal{I}^{j}_{t}$ originate from the same object sample; otherwise, they correspond to different samples.
$\tau$ is a learnable temperature parameter.

In addition, the loss $\mathcal{L}_{pair}$ is formulated as
\begin{equation}\label{eq:pairwise_triplet_sign}
\begin{aligned}
\mathcal{L}_{\mathrm{pair}}
&= 
\frac{1}{N^{+}} \sum_{\left ( i,i \right )} 
\max \left ( 0, \, \cos \left ( \mathcal{P}^{\mathcal{G},i}_{t}, \mathcal{P}^{\mathcal{G},i}_{t-1} \right ) \right )  \\
&\quad +
\frac{1}{N^{-}} \sum_{\left ( i,j \right )} \left( 1 - \cos \left ( \mathcal{P}^{\mathcal{G},i}_{t}, \mathcal{P}^{\mathcal{G},j}_{t-1} \right ) \right) ,
\end{aligned}
\end{equation}
where $N^{+}$ and $N^{-}$ denote the numbers of positive and negative pairs within each training batch, respectively.

\begin{table*}[th]
\caption{
Evaluation results on the KITTI \textit{test} set for the $Car$ category.
Superscripts $\dagger$ and $\ast$ denote the use of identical 3D object detectors, EPNet \cite{epnet} and VirConv \cite{virconv}, respectively.
``C'' and ``L'' indicate the use of camera and LiDAR modalities, respectively.
CasTrack is evaluated in the online setting for a fair comparison, whereas the remaining results are taken from the official KITTI benchmark.
The best performance value is highlighted in \textbf{bold}.
}
\label{tab1}
\centering
\renewcommand{\arraystretch}{1.1}
\scalebox{1}{
\begin{tabular}{c|c|cccccccc}
\hline
\textbf{Method} & \textbf{Input} & \textbf{HOTA} $\uparrow$ & \textbf{Det} $\uparrow$ & \textbf{AssA} $\uparrow$ & \textbf{MOTA} $\uparrow$ & \textbf{sMOTA} $\uparrow$ & \textbf{MOTP} $\uparrow$ & \textbf{IDWS} $\downarrow$ & \textbf{FRAG} $\downarrow$ \\ \hline
QD-3DT \cite{qd-3dt}           & C     & 72.77          & 74.09          & 72.19          & 85.94          & 73.29          & 85.78          & 206         & 525          \\
TripletTrack \cite{triplettrack}    & C     & 73.58          & 73.18          & 74.66          & 84.32          & 72.26          & 86.06          & 322         & 522          \\
LGM \cite{lgm}             & C     & 73.14          & 74.61          & 72.31          & 87.6           & 72.76          & 84.12          & 448         & \textbf{164} \\ \hdashline
FANTrack \cite{fantrack}        & L     & 60.85          & 64.36          & 58.69          & 75.84          & 61.49          & 82.46          & 743         & 701          \\
PolarMOT \cite{polarmot}        & L     & 75.16          & 73.94          & 76.95          & 85.08          & 71.82          & 85.63          & 462         & 599          \\
EAFFMOT \cite{EAFFMOT}         & L     & 72.28          & 71.97          & 73.08          & 84.77          & 71.56          & 85.08          & 107         & 287          \\
NC2 \cite{NC2}             & L     & 71.85          & 69.61          & 74.81          & 78.52          & 65.46          & 85.84          & 159         & 271          \\ 
CasTrack$^{\ast}$ \cite{castrack} & L     & 79.97          & 77.94          & 82.67          & 89.19          & 77.07          & 86.94          & 201         & 468          \\ \hdashline
mmMOT \cite{mmmot}           & C+L   & 62.05          & 72.29          & 54.02          & 83.23          & 70.12          & 85.03          & 733         & 570          \\
JRMOT \cite{jrmot}           & C+L   & 69.61          & 73.05          & 66.89          & 85.10          & 72.11          & 85.28          & 271         & 273          \\
JMODT$^{\dagger}$ \cite{jmodt}     & C+L   & 70.73          & 73.45          & 68.76          & 85.35          & 72.19          & 85.37          & 350         & 693          \\
EagerMOT \cite{eagermot}        & C+L   & 74.39          & 75.27          & 74.16          & 87.82          & 74.97          & 85.69          & 239         & 390          \\
DeepFusionMOT \cite{deepfusionmot}   & C+L   & 75.46          & 71.54          & 80.05          & 84.63          & 71.66          & 85.02          & 84          & 472          \\
StrongFusionMOT \cite{StrongFusionMOT} & C+L   & 75.65          & 72.08          & 79.84          & 85.53          & 72.62          & 85.07          & 58          & 416          \\
PNAS-MOT \cite{pnas-mot} & C+L   & 67.32          & 77.69          & 58.99          & 89.59          & 75.99          & 85.44          & 751          & 276          \\
BcMODT$^{\dagger}$ \cite{BcMODT}          & C+L   & 71.00          & 73.62          & 69.14          & 85.48          & 72.22          & 85.31          & 381         & 732          \\
MMF-JDT \cite{mmf-jdt}        & C+L   & 79.52          & 75.83          & 84.01          & 88.06          & 75.23          & 86.24          & 37          & 363          \\
MCCA-MOT \cite{mcca-mot}        & C+L   & 79.31          & -              & 83.49          & 86.71          & -              & \textbf{87.51} & 66          & -            \\ \hdashline
Ours$^{\dagger}$       & C+L   & 76.03          & 73.16          & 79.63          & 86.26          & 72.53          & 84.86          & 97          & 623          \\
Ours$^{\ast}$     & C+L   & \textbf{81.11} & \textbf{78.26} & \textbf{84.71} & \textbf{90.08} & \textbf{77.78} & 86.85          & \textbf{36} & 366          \\ \hline
\end{tabular}
}
\end{table*}

\section{Experiment}

\subsection{Dataset} 

We evaluate our RegTrack on the widely used KITTI \cite{kitti} and nuScenes \cite{nuscenes} benchmarks.
KITTI contains 21 training sequences with 8008 frames and 29 test sequences with 11095 frames.
nuScenes includes 850 training sequences with 34149 frames, 150 validation sequences with 6009 frames, and 150 test sequences with 6008 frames. 
In addition, KITTI uses a 64-beam LiDAR sensor that produces relatively dense point clouds, whereas nuScenes employs a 32-beam LiDAR sensor that yields sparser point clouds.


\subsection{Evaluation Metrics} \label{Evaluation Metrics}
For KITTI, we evaluate RegTrack using HOTA \cite{hota} as the main metric, which provides a comprehensive evaluation of both detection and association quality through DetA and AssA.
Additionally, we report CLEAR MOT metrics \cite{clear_mot}, including MOTA, sMOTA, MOTP, IDSW, and FRAG.
For nuScenes, we follow the official evaluation protocol, using AMOTA as the main metric by averaging recall-normalized MOTA over multiple recall thresholds, with additional reported metrics including Recall, TP, FP, FN, and IDS.




\subsection{Experimental Setup}

RegTrack is developed using Python and PyTorch, with experiments conducted on a computer equipped with an Intel Core i9 3.70GHz CPU, 64GB of RAM, and an RTX 4090 GPU. 
For the KITTI dataset, the training set is divided into two subsets: sequences \textit{0000, 0002, 0003, 0004, 0005, 0007, 0009, 0011, 0017, 0020} are used for training, while sequences \textit{0001, 0006, 0008, 0010, 0012, 0013, 0014, 0015, 0016, 0018, 0019} are used for validation. For the nuScenes dataset, 100 sequences from the training set are used for training, and the validation set is used for evaluation.

The training configuration for UTEnc includes 20 epochs, a learning rate of 0.0004, and the AdamW optimizer.
The image encoder processes input patches of size $224 \times 224$ pixels, while the point cloud encoder processes patches containing 100 points during both training and testing.


\begin{table*}[ht]
\renewcommand{\arraystretch}{1.1}
\caption{
Evaluation results on the nuScenes \textit{test} set over seven categories.
Superscripts $\dagger$ and $\ast$ denote the use of identical 3D object detectors, LargeKernel3D \cite{largekernel3d} and FocalFormer3D-F \cite{focalformer3d}, respectively.
C'' and L'' indicate the use of camera and LiDAR modalities, respectively.
Results of competing methods are taken from the official nuScenes benchmark.
The best performance value is highlighted in \textbf{bold}.
}
\label{tab2-nuscenes}
\centering
\scalebox{1}{
\begin{tabular}{c|c|ccccccccccccc}
\hline
\multirow{2}{*}{\textbf{Method}} & \multirow{2}{*}{\textbf{Input}} & \multicolumn{8}{c}{\textbf{AMOTA$\uparrow$}} & \multirow{2}{*}{\textbf{Recall$\uparrow$}} & \multirow{2}{*}{\textbf{TP$\uparrow$}} & \multirow{2}{*}{\textbf{FP$\downarrow$}} & \multirow{2}{*}{\textbf{FN$\downarrow$}} & \multirow{2}{*}{\textbf{IDS$\downarrow$}} \\ \cline{3-10}
                                 &                                 & \textbf{Overall} & Bic. & Bus  & Car  & Mot. & Ped. & Tra. & Tru. &                                  &                              &                              &                              &                               \\ \hline
StreamPETR \cite{streampetr}     & C                               & 65.3             & 56.2          & 60.6          & 75.1          & 67.5          & 72.2          & 64.6          & 61.1          & 73.3                             & 91699                        & 16637                        & 26829                        & 1037                                           \\
DualViewDistill \cite{DualViewDistill} & C                               & 66.9             & 57.3          & 63.3          & 78.8          & 70.6          & 74.3          & 67            & 56.8          & 74.1                             & 96276                        & 16499                        & 22882                        & 407                                               \\ \hdashline
SimpleTrack \cite{simpletrack}   & L                               & 66.8             & 40.7          & 71.5          & 82.3          & 67.4          & 79.6          & 67.3          & 58.7          & 70.3                             & 95539                        & 17514                        & 24351                        & 575                                           \\
3DMOTFormer \cite{3dmotformer}   & L                               & 68.2             & 37.4          & 74.9          & 82.1          & 70.5          & 80.7          & 69.6          & 62.6          & 69.3                             & 95790                        & 18322                        & 23337                        & 438                                                 \\
VoxelNeXt \cite{voxelnext}       & L                               & 71.0             & 52.6          & 74.7          & 82.6          & 73.1          & 76.0          & 73.8          & 64.4          & 76.5                             & 97075                        & 18348                        & 21836                        & 654                                           \\
ShaSTA \cite{shasta}            & L                               & 69.6             & 41.0          & 73.3          & 83.8          & 72.7          & 81            & 70.4          & 65.0          & 71                               & 97799                        & 16746                        & 21293                        & 473                                        \\
NEBP-V3 \cite{nebp}             & L                               & 74.6             & 49.9          & 78.6          & 86.1          & 80.7          & 83.4          & 76.1          & 67.3          & 76.3                             & 99872                        & 17243                        & 19316                        & 377                                           \\
Poly-MOT$^{\dagger}$ \cite{poly-mot}   & L                               & 75.4             & 58.2          & 78.6          & 86.5          & 81.0          & 82.0          & 75.1          & 66.2          & 78.3                             & 101317                       & 19673                        & 17956                        & 292                                     \\
Fast-Poly$^{\dagger}$ \cite{fast-poly} & L                               & 75.8             & 57.3          & 76.7          & 86.3          & 82.6 & 85.2          & 76.8          & 65.6          & 77.6                             & 100824                       & 17098                        & 18415                        & 326                                         \\ \hdashline
TransFusion \cite{transfusion}   & C+L                             & 71.8             & 53.9          & 75.4          & 82.1          & 72.1          & 79.6          & 73.1          & 66.3          & 75.8                             & 96775                        & 16232                        & 21846                        & 944                                    \\
UVTR \cite{uvtr}                 & C+L                             & 70.1             & 51.7          & 73.7          & 82.5          & 69.5          & 80.9          & 72.8          & 59.3          & 75                               & 98434                        & 15615                        & 20190                        & 941                                              \\
EagerMOT \cite{eagermot}         & C+L                             & 67.7             & 58.3          & 74.0          & 81.0          & 62.5          & 74.4          & 63.6          & 59.7          & 72.7                             & 93484                        & 17705                        & 24925                        & 1156                                             \\
DeepFusionMOT \cite{deepfusionmot} & C+L                             & 63.5             & 52.0          & 70.8          & 72.5          & 69.6          & 55.4          & 64.9          & 59.6          & 69.6                             & 84304                        & 19303                        & 33556                        & 1075                                               \\
MotionTrack \cite{motiontrack}  & C+L                             & 55.0             & -             & -             & -             & -             & -             & -             & -             & -                                & -                            & -                            & -                            & 8716                                          \\
FocalFormer3D-F$^{\ast}$ \cite{focalformer3d} & C+L                             & 73.9             & 54.1          & \textbf{79.2} & 83.4          & 74.6          & 84.1          & 75.2          & 66.9          & 75.9                             & 97987                        & 15547                        & 20754                        & 824                                  \\
MSMDFusion \cite{msmdfusion}   & C+L                             & 74.0             & 57.4          & 76.7          & 84.9          & 75.4          & 80.7          & 75.4          & 67.1          & 76.3                             & 98624                        & \textbf{14789}               & 19853                        & 1088                                      \\
BEVFusion \cite{bevfusion}    & C+L                             & 74.1             & 56.0          & 77.9          & 83.1          & 74.8          & 83.7          & 73.4          & \textbf{69.5} & 77.9                             & 99664                        & 19997                        & 19395                        & 506                              \\
CAMO-MOT \cite{camo-mot}      & C+L                             & 75.3             & 59.2 & 77.7          & 85.8          & 78.2          & 85.8 & 72.3          & 67.7          & 79.1                    & 101049                       & 17269                        & 18192                        & 324               \\
FutrTrack$^{\ast}$ \cite{futrtrack}    & C+L                             & 74.7             & 55.8          & 79.0          & 83.8          & 76.8          & 84.2          & 75.4          & 67.8          & 78.0                             & 99204                        & 18295                        & 19914                        & 447                                                       \\
Xie et al.$^{\ast}$ \cite{xie2025multi}    & C+L                             & 74.0             & -          & -          & -          & -          & -          & -          & -          & 76.1                             & -                        & 16289                        & -                        & -                                                       \\ \hdashline
Ours$^{\dagger}$                     & C+L                             & 76.5    & 58.5   & 77.6  & 87.2  & 81.9   & 85.7          & \textbf{78.6} & 66.0     & 77.8   & 101980   & 16611   & 17319               & \textbf{266}                                    \\
Ours$^{\ast}$                     & C+L                             & \textbf{78.0}    & \textbf{63.1}          & 78.7          & \textbf{88.0} & \textbf{83.2}          & \textbf{86.8}          & 77.6 & 68.3          & \textbf{79.9}                             & \textbf{103868}              & 15827                        & \textbf{15378}               & 319                                     \\
\hline
\end{tabular}
}
\end{table*}

\subsection{Main Results}

\subsubsection{Quantitative Evaluation on KITTI} 

Table~\ref{tab1} reports comprehensive experiments on the KITTI test set.
The results indicate that, when using the same 3D detections from EPNet \cite{epnet} or VirConv \cite{virconv} as inputs, RegTrack significantly outperforms JMODT \cite{jmodt}, BcMODT \cite{BcMODT}, and CasTrack \cite{castrack}, with particularly notable improvements on the HOTA metric.
Specifically, when utilizing 3D detections from EPNet \cite{epnet}, RegTrack surpasses JMODT and BcMODT by 5.30\% and 5.03\% in HOTA, 10.87\% and 10.49\% in AssA, and 0.91\% and 0.78\% in MOTA, respectively, while also reducing ID switches by 253 and 284 and FRAG by 70 and 109.
Similarly, when using 3D detections from VirConv, RegTrack achieves improvements of 1.14\% in HOTA, 2.04\% in AssA, 0.89\% in MOTA, and 0.71\% in sMOTA, along with reductions of 165 ID switches and 102 FRAG compared with CasTrack.
In addition, RegTrack outperforms 15 other competitors.
Collectively, these results demonstrate the state-of-the-art performance of our method.

\subsubsection{Quantitative Evaluation on nuScenes} 
To evaluate the generalizability of RegTrack across diverse object categories and its robustness to sparse point clouds, we conduct additional experiments on the more challenging nuScenes test set, covering seven object categories, as summarized in Table \ref{tab2-nuscenes}.
The results show that RegTrack outperforms 19 competing methods.
Specifically, when using the same 3D detections from LargeKernel3D \cite{largekernel3d}, RegTrack improves over Poly-MOT \cite{poly-mot} by 1.1\% in AMOTA and 663 in TP, and over Fast-Poly \cite{fast-poly} by 0.7\% in AMOTA and 1156 in TP. Moreover, RegTrack reduces FP, FN, and IDS by 3062, 637, and 26 compared with Poly-MOT, and by 487, 637, and 60 compared with Fast-Poly.
Similarly, when using the same 3D detections from FocalFormer3D-F \cite{focalformer3d}, RegTrack improves over FocalFormer3D-F by 4.1\% in AMOTA, 4.0\% in Recall, and 5881 in TP, and over FutrTrack \cite{futrtrack} by 3.3\%, 1.9\%, and 4664, respectively. In addition, RegTrack reduces FP, FN, and IDS by 1064, 5376, and 505 compared with FocalFormer3D-F, and by 2468, 4536, and 128 compared with FutrTrack.
Furthermore, under the AMOTA metric across all seven categories, when using 3D detections from FocalFormer3D-F, RegTrack achieves the highest scores on the \textit{bicycle}, \textit{car}, \textit{motorcycle}, and \textit{pedestrian} categories among the 19 competing methods.
Overall, these results demonstrate that our method achieves state-of-the-art robustness across diverse object categories, even under sparse point cloud conditions.

\begin{figure*}[ht]
    \centering
    \includegraphics[width=\linewidth]{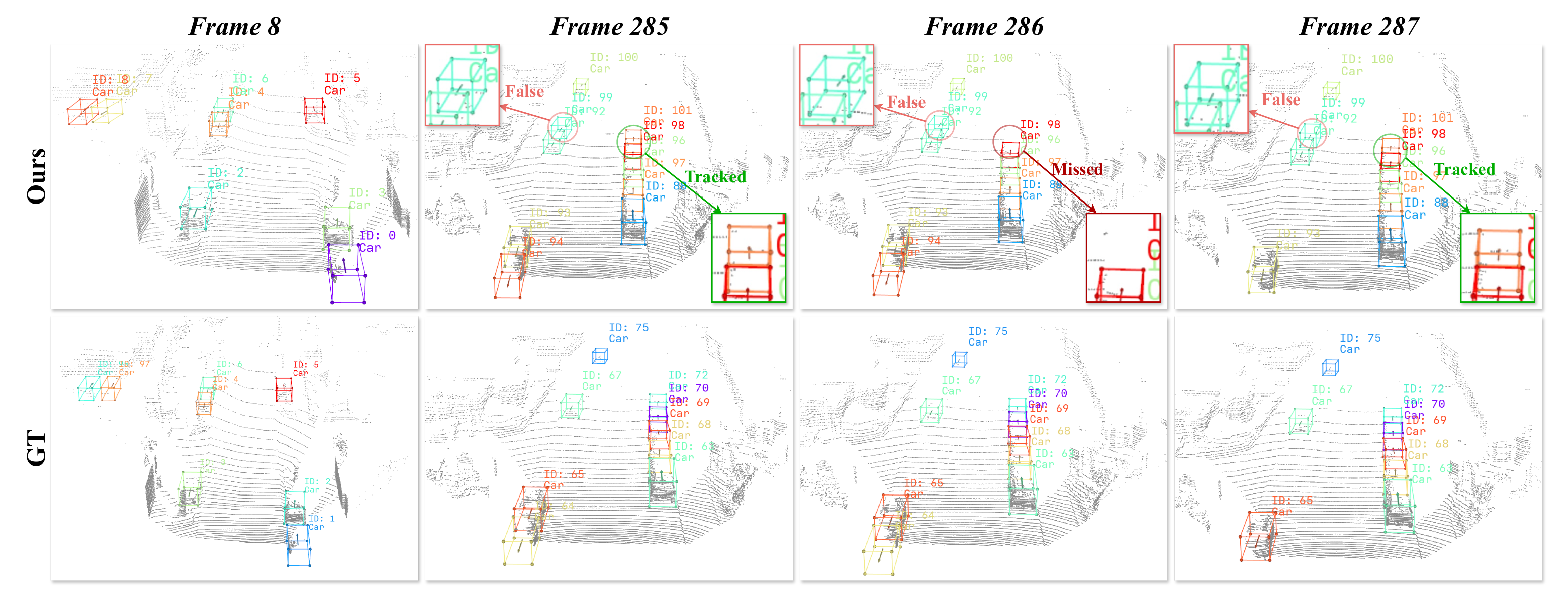}
    \caption{\textbf{Qualitative evaluation of our RegTrack on the KITTI \textit{validation} set.} 
    It visualizes the 3D MOT results on sequence \textit{0001}, using 3D detections generated by VirConv \cite{virconv} as input, where each column corresponds to a single frame.
    }
    \label{fig:vis_RegTrack_kitti}
\end{figure*}

\begin{table*}[th]
\caption{
Ablation study on the impact of using point cloud, image, and geometry cues during training for RegTrack on the nuScenes \textit{validation} set over seven categories.
3D detections are obtained using CenterPoint \cite{centerpoint}.
}
\label{tab3}
\centering
\renewcommand{\arraystretch}{1.2}
\scalebox{1}{
\begin{tabular}{cc|c|cccccc}
\hline
\multicolumn{2}{c|}{\textbf{UTEnc}} & \textbf{MoE-GEnc} & \multirow{2}{*}{\textbf{AMOTA}$\uparrow$} & \multirow{2}{*}{\textbf{Recall}$\uparrow$} & \multirow{2}{*}{\textbf{TP}$\uparrow$} & \multirow{2}{*}{\textbf{FP}$\downarrow$} & \multirow{2}{*}{\textbf{FN}$\downarrow$} & \multirow{2}{*}{\textbf{IDS}$\downarrow$} \\ \cline{1-3}
Point Cloud     & Image    & Geometry &                        &                         &                     &                     &                     &                      \\ \hline
\ding{52}      &          &                           & 70.0                  & 73.8                   & 81870               & 14607               & 19614               & 413                  \\
\ding{52}      & \ding{52} &                          & 72.0                  & 74.2                   & 82243               & 12115               & 19235               & 419                  \\
\ding{52}      & \ding{52} & \ding{52}                & \textbf{73.8}         & \textbf{75.1}          & \textbf{83908}      & \textbf{11380}      & \textbf{17718}      & \textbf{271}         \\ \hline
\end{tabular}
}
\end{table*}

\begin{table}[th]
\caption{
Ablation study of RegTrack with and without image cues during inference on the nuScenes \textit{validation} set across seven categories.
3D detections are obtained using CenterPoint \cite{centerpoint}.
}
\label{tab3_2}
\centering
\renewcommand{\arraystretch}{1.2}
\scalebox{1}{
\begin{tabular}{c|cccccc}
\hline
\textbf{Image} & \textbf{AMOTA}$\uparrow$ & \textbf{Recall}$\uparrow$ & \textbf{TP}$\uparrow$ & \textbf{FP}$\downarrow$ & \textbf{FN}$\downarrow$ & \textbf{IDS}$\downarrow$ \\ \hline
\ding{52}              & 73.5          & \textbf{75.5}           & \textbf{84075}       & 11737       & \textbf{17486}       & 336          \\
\ding{55}              & \textbf{73.8}          & 75.1           & 83908       & \textbf{11380}       & 17718       & \textbf{271}          \\ \hline
\end{tabular}
}
\end{table}

\begin{table*}[th]
\caption{
Ablation study on the positional encoding type and channel dimensionality for input point clouds in RegTrack on the nuScenes \textit{validation} set over seven categories. 3D detections are obtained using CenterPoint \cite{centerpoint}.
}
\label{tab4}
\centering
\renewcommand{\arraystretch}{1.2}
\scalebox{1}{
\begin{tabular}{cc|cc|cccccc}
\hline
\multicolumn{2}{c|}{\textbf{Positional Encoding}} & \multicolumn{2}{c|}{\textbf{Input Channels}} & \multirow{2}{*}{\textbf{AMOTA}$\uparrow$} & \multirow{2}{*}{\textbf{Recall}$\uparrow$} & \multirow{2}{*}{\textbf{TP}$\uparrow$} & \multirow{2}{*}{\textbf{FP}$\downarrow$} & \multirow{2}{*}{\textbf{FN}$\downarrow$} & \multirow{2}{*}{\textbf{IDS}$\downarrow$} \\ \cline{1-4}
MLP-based          & Sinusoidal         & (x, y, z)         & (x, y)        &                        &                         &                     &                     &                     &                      \\ \hline
\ding{52}                     &                       & \ding{52}                  &                & 71.1                  & 73.8                   & 81907               & 13575               & 19591               & 399                  \\
                      & \ding{52}                     & \ding{52}                  &                & 73.6                  & 75.0                   & 83882               & 11467               &     \textbf{17671}      & 344                  \\
                      & \ding{52}                     &                    & \ding{52}              & \textbf{73.8}         & \textbf{75.1}          & \textbf{83908}      & \textbf{11380}      & 17718               & \textbf{271}         \\ \hline
\end{tabular}
}
\end{table*}

\begin{table}[th]
\caption{
Ablation study on the number of experts in MoE-GEnc for RegTrack on the nuScenes \textit{validation} set over seven categories. 3D detections are obtained using CenterPoint \cite{centerpoint}.
}
\label{tab5}
\centering
\renewcommand{\arraystretch}{1.2}
\scalebox{0.93}{
\begin{tabular}{ccc|cccccc}
\hline
\multicolumn{3}{c|}{\textbf{Number}} & \multirow{2}{*}{\textbf{AMOTA}$\uparrow$} & \multirow{2}{*}{\textbf{Recall}$\uparrow$} & \multirow{2}{*}{\textbf{TP}$\uparrow$} & \multirow{2}{*}{\textbf{FP}$\downarrow$} & \multirow{2}{*}{\textbf{FN}$\downarrow$} & \multirow{2}{*}{\textbf{IDS}$\downarrow$} \\ \cline{1-3}
1       & 2       & 3       &                        &                         &                     &                     &                     &                      \\ \hline
\ding{52}       &         &         & 73.3                  & 74.8                   & 83885               & 11455               & 17730               & 282            \\
        & \ding{52}       &         & \textbf{73.8}         & 75.1                   & 83908               & \textbf{11380}      & 17718               & \textbf{271}   \\
        &         & \ding{52}       & 73.6                  & \textbf{75.4}          & \textbf{83938}      & 11389               & \textbf{17678}      & 281            \\ \hline
\end{tabular}
}
\end{table}

\begin{table}[th]
\caption{
Ablation study on local features (LF), global features (GF), and hybrid attention (HA) in LG-PEnc on the nuScenes \textit{validation} set over seven categories. 3D detections are obtained using CenterPoint \cite{centerpoint}.
}
\label{tab6}
\centering
\renewcommand{\arraystretch}{1.2}
\scalebox{0.9}{
\begin{tabular}{ccc|cccccc}
\hline
\multicolumn{3}{c|}{\textbf{LG-PEnc}} & \multirow{2}{*}{\textbf{AMOTA}$\uparrow$} & \multirow{2}{*}{\textbf{Recall}$\uparrow$} & \multirow{2}{*}{\textbf{TP}$\uparrow$} & \multirow{2}{*}{\textbf{FP}$\downarrow$} & \multirow{2}{*}{\textbf{FN}$\downarrow$} & \multirow{2}{*}{\textbf{IDS}$\downarrow$} \\ \cline{1-3}
LF         & GF         & HA         &                                 &                                  &                              &                              &                              &                               \\ \hline
\ding{52}          &            &            & 73.1                           & 74.5                            & 83857                        & 11727                        & 17719                        & 321                           \\
\ding{52}          & \ding{52}          &            & 73.6                           & \textbf{75.2}                            & \textbf{83935}                        & 11626                        & \textbf{17668}                        & 294                           \\
\ding{52}          & \ding{52}          & \ding{52}          & \textbf{73.8}                           & 75.1                            & 83908                        & \textbf{11380}                        & 17718                        & \textbf{271}                           \\ \hline
\end{tabular}
}
\end{table}

\begin{table*}[th]
\caption{
Ablation study of LG-PEnc on the nuScenes validation set over seven categories, comparing point-wise vs. channel-wise attention normalization for local features, and fixed average vs. adaptive weighted aggregation for local-global features.
3D detections are obtained using CenterPoint \cite{centerpoint}.
}
\label{tab6-2}
\centering
\renewcommand{\arraystretch}{1.2}
\scalebox{1}{
\begin{tabular}{cc|cc|cccccc}
\hline
\multicolumn{2}{c|}{\textbf{Normalization}} & \multicolumn{2}{c|}{\textbf{Aggregation}} & \multirow{2}{*}{\textbf{AMOTA}$\uparrow$} & \multirow{2}{*}{\textbf{Recall}$\uparrow$} & \multirow{2}{*}{\textbf{TP}$\uparrow$} & \multirow{2}{*}{\textbf{FP}$\downarrow$} & \multirow{2}{*}{\textbf{FN}$\downarrow$} & \multirow{2}{*}{\textbf{IDS}$\downarrow$} \\ \cline{1-4}
Point-wise               & Channel-wise               & Fixed         & Adaptive         &                                 &                                  &                              &                              &                              &                               \\ \hline
\ding{52}                   &                       & \ding{52}                     &                   & 73.1                            & 74.5                             & 83825                        & 11801                        & 17756                        & 316                           \\
                    & \ding{52}                     & \ding{52}                     &                   & 73.0                            & \textbf{75.4}                    & \textbf{84035}               & 12037                        & \textbf{17525}               & 337                           \\
                    & \ding{52}                     &                       & \ding{52}                 & \textbf{73.8}                   & 75.1                             & 83908                        & \textbf{11380}               & 17718                        & \textbf{271}                  \\ \hline
\end{tabular}
}
\end{table*}

\begin{table*}[th]
\caption{
Evaluation results of our LG-PEnc compared with other point cloud encoders on the nuScenes validation set over seven categories. 3D detections are obtained using CenterPoint \cite{centerpoint}.
}
\label{tab7}
\centering
\renewcommand{\arraystretch}{1.2}
\scalebox{1}{
\begin{tabular}{cll|cccccccccc|cc}
\hline
\multicolumn{3}{c|}{\multirow{2}{*}{\textbf{Point Cloud Encoder}}} & \multicolumn{8}{c}{\textbf{AMOTA}$\uparrow$}                                                                                             & \multirow{2}{*}{\textbf{TP}$\uparrow$} & \multirow{2}{*}{\textbf{IDS}$\downarrow$} & \multirow{2}{*}{\textbf{Params} (M)} & \multirow{2}{*}{\textbf{FLOPs} (GFLOPs)} \\ \cline{4-11}
\multicolumn{3}{c|}{}                                              & \textbf{Overall} & Bic.        & Bus           & Car           & Mot.          & Ped.          & Tra.          & Tru.          &                              &                               &                                      &                                          \\ \hline
\multicolumn{3}{c|}{PointNet \cite{pointnet}}                                      & 72.1             & 56.1        & 88.3          & 83.1          & 72.1          & 81.8          & 53.5          & 69.6          & 83706                        & 357                           & 4.08                                 & 0.1                                      \\
\multicolumn{3}{c|}{PointNet++(SSG) \cite{pointnet++}}                               & 73.0             & 55.7        & 88.2          & 85.2          & 75.4          & 82.5          & 53.3          & 70.9          & 84031                        & 475                           & 1.8                                  & 0.07                                     \\
\multicolumn{3}{c|}{PointNet++(MSG) \cite{pointnet++}}                               & 72.7             & 56.0          & 87.4          & 83.7          & 74.9          & 82.0            & 54.0            & 70.6          & \textbf{84191}               & 609                           & 2.07                                 & 0.25                                     \\
\multicolumn{3}{c|}{PointConv \cite{pointconv}}                                     & 73.0               & 56.1        & 88.3          & 85.4          & 75.1          & 82.1          & 53.0            & 71.2          & 84085                        & 432                           & 19.9                                 & 0.19                                     \\
\multicolumn{3}{c|}{DGCNN \cite{dgcnn}}                                         & 72.9             & 56.9        & 87.8          & 85.1          & 73.7          & 81.4          & 53.9          & 71.2          & 84110                        & 555                           & 0.66                                 & 0.16                                     \\
\multicolumn{3}{c|}{PointMixer \cite{pointmixer}}                                    & 72.1             & 55.5        & 88.3          & 85.0            & 70.3          & 81.7          & 53.2          & 70.7          & 84043                        & 513                           & 4.47                                 & 0.15                                     \\ \hdashline
\multicolumn{3}{c|}{Ours (LG-PEnc)}                                 & \textbf{73.8}    & \textbf{57} & \textbf{88.4} & \textbf{85.7} & \textbf{76.7} & \textbf{83.2} & \textbf{54.2} & \textbf{71.2} & 83908                        & \textbf{271}                  & 2.6                                  & 0.32                                     \\ \hline
\end{tabular}
}
\end{table*}

\begin{table}[th]
\caption{
Model complexity analysis. Note that point cloud and geometry inputs use paired object samples, whereas image inputs use a single anchor-aligned object sample.
}
\label{tab8}
\centering
\renewcommand{\arraystretch}{1.2}
\scalebox{0.95}{
\begin{tabular}{cc|c|cc}
\hline
\multicolumn{2}{c|}{\textbf{UTEnc}} & \textbf{MoE-GEnc} & \multirow{2}{*}{\textbf{Params} (M)} & \multirow{2}{*}{\textbf{FLOPs} (GFLOPs)} \\ \cline{1-3}
Point Cloud         & Image         & Geometry          &                             &                                 \\ \hline
\ding{52}           &               &                   & 2.39                        & 0.32                         \\
\ding{52}           &               & \ding{52}         & 2.60                        & 0.32                         \\
\ding{52}           & \ding{52}     & \ding{52}         & 89.19                       & 35.49                        \\ \hline
\end{tabular}
}
\end{table}

\begin{table*}[th]
\caption{
Runtime performance analysis on the KITTI and nuScenes \textit{test} sets, including GPU and CPU memory usage and FPS. 
VirConv \cite{virconv} is used for KITTI and FocalFormer3D-F \cite{focalformer3d} for nuScenes.
Data loading and object-level point cloud cropping and resampling are excluded from the reported computation.
On nuScenes, incorporating image information exceeds available GPU memory due to the large number of objects per scene.
}
\label{tab9}
\centering
\renewcommand{\arraystretch}{1.2}
\scalebox{1}{
\begin{tabular}{cc|c|ccc|ccc}
\hline
\multicolumn{2}{c|}{\textbf{UTEnc}} & \textbf{MoE-GEnc} & \multicolumn{3}{c|}{\textbf{KITTI}}                  & \multicolumn{3}{c}{\textbf{nuScenes}}                \\ \hline
Point Cloud         & Image         & Geometry          & \textbf{GPU} (MB) & \textbf{CPU} (MB) & \textbf{FPS} & \textbf{GPU} (MB) & \textbf{CPU} (MB) & \textbf{FPS} \\ \hline
\ding{52}                   &               &                   & 248               & 122               & 245          & 1011              & 138               & 27           \\
\ding{52}                   &               & \ding{52}         & 248               & 132               & 156          & 1165              & 153               & 19           \\
\ding{52}                   & \ding{52}     & \ding{52}         & 6990              & 143               & 29           & -              & -               & -            \\ \hline
\end{tabular}
}
\end{table*}

\begin{table*}[th]
\caption{
Generalizability analysis of the association threshold on the KITTI \textit{training} set (21 sequences) and the nuScenes \textit{validation} set, using VirConv \cite{virconv} for KITTI and CenterPoint \cite{centerpoint} for nuScenes.
}
\label{tab10}
\centering
\renewcommand{\arraystretch}{1.2}
\scalebox{1}{
\begin{tabular}{c|cccc|cccccccccc}
\hline
\multirow{3}{*}{\textbf{Thresholds}} & \multicolumn{4}{c|}{\textbf{KITTI}}                              & \multicolumn{10}{c}{\textbf{nuScenes}}                                                                                                                                                        \\ \cline{2-15} 
                                     & \textbf{HOTA}$\uparrow$  & \textbf{MOTA}$\uparrow$  & \textbf{sMOTA}$\uparrow$ & \textbf{IDWS}$\downarrow$ & \multicolumn{8}{c}{\textbf{AMOTA}$\uparrow$}                                                                                             & \multirow{2}{*}{\textbf{TP}$\uparrow$} & \multirow{2}{*}{\textbf{IDS}$\downarrow$} \\ \cline{2-13}
                                     & \multicolumn{4}{c|}{Car}                                         & \textbf{Overall} & Bic.        & Bus           & Car           & Mot.          & Ped.          & Tra.          & Tru.          &                              &                               \\ \hline
0.1                                  & 83.00          & \textbf{86.11} & \textbf{78.56} & 20            & 72.2             & 55.6        & 85.4          & 83.8          & 73.3          & 82.8          & 54            & 70.4          & 83978                        & 640                           \\
0.2                                  & 82.97          & 86.09          & 78.54          & 19            & 72.5             & 55.6        & 85.6          & 83.9          & 74.6          & 82.8          & 54.2          & 70.7          & 83920                        & 448                           \\
0.3                                  & \textbf{83.02} & \textbf{86.11} & 78.55          & 16            & 73.3             & 55.8        & 88.4          & 85.6          & 75.1          & 82.7          & 54.3          & 71.1          & 83890                        & 343                           \\
0.4                                  & \textbf{83.02} & \textbf{86.11} & \textbf{78.56} & \textbf{15}   & 73.6             & 56.9        & 88.3          & \textbf{85.7} & 75.5          & 83.1          & \textbf{54.7} & 71.1          & \textbf{84608}               & 310                           \\ 
\textbf{0.5}                         & 83.01          & 86.10          & 78.54          & 17            & \textbf{73.8}    & \textbf{57.0} & \textbf{88.4} & \textbf{85.7} & \textbf{76.7} & \textbf{83.2} & 54.2          & \textbf{71.2} & 83908                        & \textbf{271}                  \\
0.6                                  & 83.01          & 86.09          & 78.54          & 17            & 73.5             & 56.9        & 88.3          & 85.5          & 76.6          & 82.7          & 53.7          & \textbf{71.2} & 83869                        & 277                           \\
0.7                                  & 83.00          & 86.07          & 78.52          & 17            & 73.1             & 56.9        & 88.3          & 84.8          & 75.7          & 82.0          & 53.6          & 70.4          & 83887                        & 289                           \\
0.8                                  & 82.91          & 85.98          & 78.43          & 18            & 71.0             & 54.7        & 88.2          & 81.9          & 73.4          & 79.6          & 51.9          & 67.1          & 81809                        & 312                           \\
0.9                                  & 82.22          & 85.73          & 78.16          & 34            & 64.5             & 47.5        & 84.7          & 75.3          & 66.1          & 72.5          & 45.8          & 59.8          & 75734                        & 353                           \\ \hline
\end{tabular}
}
\end{table*}

\begin{table*}[th]
\caption{
Generalizability analysis of the association threshold in the advanced competitor Poly-MOT~\cite{poly-mot} on the nuScenes \textit{validation} set over seven categories, using CenterPoint~\cite{centerpoint} as the 3D detector. The best performance is achieved with the class-specific thresholds reported in the Poly-MOT paper: 1.34 for \textit{bicycle}, 1.55 for \textit{bus}, 1.31 for \textit{car}, 1.44 for \textit{motorcycle}, 1.69 for \textit{pedestrian}, 1.25 for \textit{trailer}, and 1.21 for \textit{truck}.
}
\label{tab10-2}
\centering
\renewcommand{\arraystretch}{1.2}
\scalebox{1}{
\begin{tabular}{c|ccccccccccccc}
\hline
\multirow{2}{*}{\textbf{Thresholds}} & \multicolumn{8}{c}{\textbf{AMOTA}}                                                                                               & \multirow{2}{*}{\textbf{Recall}} & \multirow{2}{*}{\textbf{TP}} & \multirow{2}{*}{\textbf{FP}} & \multirow{2}{*}{\textbf{FN}} & \multirow{2}{*}{\textbf{IDS}} \\ \cline{2-9}
                                     & \textbf{Overall} & Bic.          & Bus           & Car           & Mot.          & Ped.          & Tra.          & Tru.          &                                  &                              &                              &                              &                               \\ \hline
\textbf{Best}                        & \textbf{73.1}    & 54.5          & 87.3          & 86.3          & 78.3          & \textbf{83.0} & 51.0          & 71.2          & 74.7                             & 84072                        & 13051                        & 17593                        & \textbf{232}                  \\ \hdashline
1.0                                  & 69.7             & 47.0          & 85.5          & 86.3          & 69.3          & 78.1          & 50.8          & 70.7          & 74.6                             & 84915                        & 14877                        & \textbf{16181}               & 801                           \\
1.1                                  & 70.6             & 50.6          & 85.8          & \textbf{86.4} & 70.3          & 79.0          & 50.8          & 71.2          & \textbf{75.5}                    & \textbf{85023}               & 14619                        & 16225                        & 649                           \\
1.2                                  & 72.2             & 54.4          & \textbf{87.5} & \textbf{86.4} & 74.9          & 80.2          & 50.7          & 71.0          & 74.6                             & 83980                        & 13331                        & 17407                        & 510                           \\
1.3                                  & 72.3             & 54.5          & 87.2          & \textbf{86.4} & 74.8          & 81.4          & 50.9          & \textbf{71.3} & 74.4                             & 83389                        & 12675                        & 18143                        & 365                           \\
1.4                                  & 72.6             & 54.3          & 87.1          & 86.1          & 77.1          & 81.6          & \textbf{51.1} & 70.6          & 74.1                             & 82946                        & \textbf{12385}               & 18670                        & 281                           \\
1.5                                  & 72.6             & 54.6          & 87.1          & 85.8          & \textbf{78.4} & 81.8          & 50.8          & 69.8          & 72.8                             & 83242                        & 13023                        & 18380                        & 275                           \\
1.6                                  & 72.3             & \textbf{54.7} & 86.9          & 84.1          & 78.2          & 82.5          & 50.3          & 69.6          & 73.1                             & 83733                        & 13903                        & 17876                        & 288                           \\
1.7                                  & 71.7             & 54.5          & 86.4          & 83.0          & 76.9          & \textbf{83.0} & 49.7          & 68.0          & 73.4                             & 83616                        & 15324                        & 17979                        & 302                           \\
1.8                                  & 69.4             & 54.0          & 83.4          & 79.4          & 75.0          & 82.2          & 48.1          & 64.0          & 70.4                             & 81565                        & 16671                        & 19965                        & 367                           \\
1.9                                  & 63.0             & 50.8          & 74.7          & 74.2          & 68.7          & 80.4          & 39.3          & 52.7          & 65.9                             & 76791                        & 18381                        & 24692                        & 414                           \\ \hline
\end{tabular}
}
\end{table*}


\begin{table*}[th]
\caption{
Cross-domain generalizability analysis between the KITTI \textit{training} set (21 sequences) and the nuScenes \textit{validation} set. \textit{Source} and \textit{Target} indicate the datasets used for training and evaluation. VirConv \cite{virconv} is adopted for KITTI, and CenterPoint \cite{centerpoint} is used for nuScenes.
}
\label{tab11}
\centering
\renewcommand{\arraystretch}{1.2}
\scalebox{1}{
\begin{tabular}{c|cccccc}
\hline
\textbf{Source $\rightarrow$ Target} & \textbf{AMOTA}$\uparrow$ & \textbf{Recall}$\uparrow$ & \textbf{TP}$\uparrow$   & \textbf{FP}$\downarrow$   & \textbf{FN}$\downarrow$    & \textbf{IDS}$\downarrow$  \\ \hline
KITTI $\rightarrow$ nuScenes         & 73.2           & 74.7            & 83352         & 11627         & 18253          & 292           \\
nuScenes $\rightarrow$ nuScenes      & 73.8           & 75.1            & 83908         & 11380         & 17718          & 271           \\ \hdashline
$\Delta$                                & $+$0.6            & $+$0.4             & $+$556           & $-$247          & $-$535           & $-$21           \\ \hline \hline
\textbf{Source $\rightarrow$ Target} & \textbf{HOTA}$\uparrow$  & \textbf{Det}$\uparrow$    & \textbf{AssA}$\uparrow$ & \textbf{MOTA}$\uparrow$ & \textbf{sMOTA}$\uparrow$ & \textbf{IDWS}$\downarrow$ \\ \hline
nuScenes $\rightarrow$ KITTI         & 83.02          & 80.08           & 86.21         & 86.08         & 78.53          & 19            \\
KITTI $\rightarrow$ KITTI            & 83.01          & 80.09           & 86.19         & 86.10         & 78.54          & 17            \\ \hdashline
$\Delta$                                & $-$0.01          & $+$0.01            & $-$0.02         & $+$0.02          & $+$0.01           & $-$2         \\ \hline
\end{tabular}
}
\end{table*}



\subsubsection{Qualitative Evaluation} 

To complement the quantitative analysis, we conduct a qualitative evaluation on the KITTI validation set. Fig. \ref{fig:vis_RegTrack_kitti} visualizes the 3D MOT performance of RegTrack, using 3D detections generated by VirConv \cite{virconv} as input. RegTrack demonstrates excellent performance, producing robust and accurate 3D trajectories while effectively suppressing false trajectories.
Furthermore, columns 2 to 4 in Fig. \ref{fig:vis_RegTrack_kitti} show that RegTrack can mitigate ID switches caused by temporarily missed detections, as exemplified by the object with ID 101.

\subsection{Ablation Study}
To evaluate the effectiveness of UTEnc and MoE-GEnc, the core components of RegTrack, we perform extensive ablation studies on the nuScenes validation set across seven object categories, as reported in Table \ref{tab3} and \ref{tab3_2}. 
We further investigate the impact of key architectural design choices on 3D MOT performance, including the type of positional encoding (Table \ref{tab4}), the dimensionality of point cloud input channels (Table \ref{tab4}), and the number of experts in MoE-GEnc (Table \ref{tab5}).
In addition, we analyze the contributions of local features, global features, and the hybrid attention mechanism in LG-PEnc (see Table~\ref{tab6}), as well as the effects of different attention normalization strategies for local features and different aggregation schemes for local and global features (see Table~\ref{tab6-2}).
We also evaluate the effect of the proposed LG-PEnc on 3D MOT performance in comparison with alternative point cloud encoders (Table \ref{tab7}).

\subsubsection{Effects of UTEnc and MoE-GEnc} 
Under the training supervision of the tri-cue unification loss (see Eq. \ref{eq:tu loss}), UTEnc equipped with MoE-GEnc can jointly learn information from point cloud, image, and geometric cues to produce discriminative object representations for constructing the robust association metric.
Table \ref{tab3} reports the impact of point cloud and image cues within UTEnc, as well as the geometric cues learned by MoE-GEnc, on 3D MOT performance evaluated on the nuScenes validation set.
When UTEnc uses only point clouds, it achieves an AMOTA of 69.9\%, Recall of 73.6\%, TP of 81913, FP of 15224, FN of 19552, and IDS of 432. When UTEnc incorporates both point cloud and image cues while still excluding MoE-GEnc, AMOTA, Recall, and TP increase by 1.7\%, 0.9\%, and 537, respectively, while FP, FN, and IDS decrease by 2602, 520, and 17, respectively.
With the further inclusion of MoE-GEnc for learning geometric cues, RegTrack achieves its best overall performance, reaching an AMOTA of 73.8\%. Specifically, AMOTA, Recall, and TP increase by 2.2\%, 0.6\%, and 1458, respectively, while TP, FN, and IDS decrease by 1241, 1314, and 144, respectively.
These results indicate that all three cues contribute positively to our method.

\begin{figure}[h]
    \centering
    \includegraphics[width=1\linewidth]{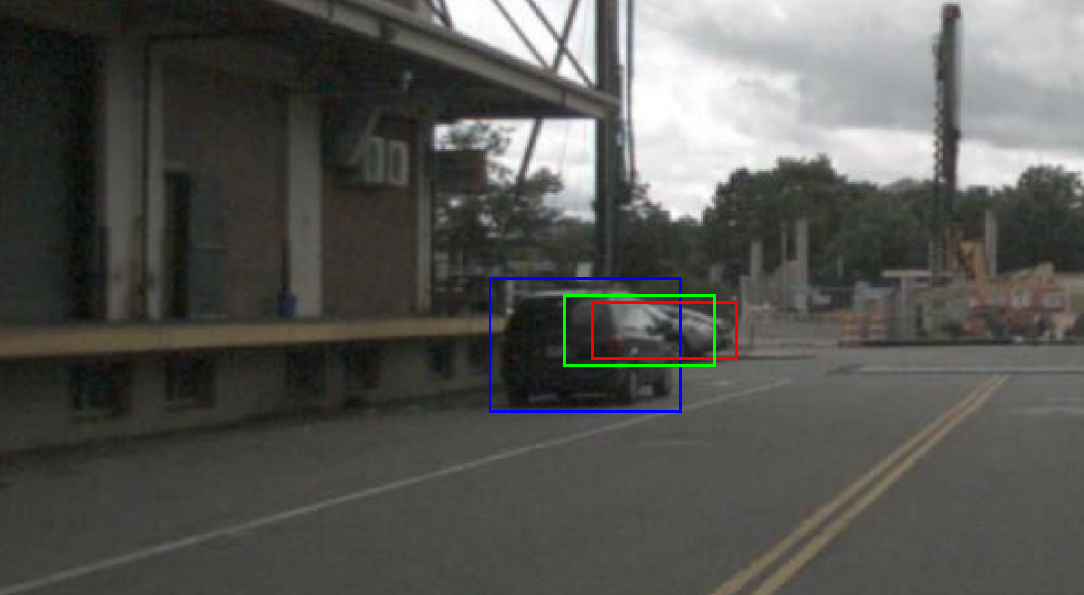}
    \caption{
    \textbf{Examples of mutual object occlusion under low-light conditions.} 
    The two objects enclosed by the yellow and red boxes are almost entirely occluded by the object highlighted in the blue box.
    }
    \label{fig:camera_cases}
\end{figure}

In addition, Table~\ref{tab3_2} analyzes the impact of incorporating image information during inference. 
Specifically, image features extracted by the CLIP encoder are projected into the embedding space via a trainable projection layer and concatenated with point cloud features along the channel dimension. The fused representation is then processed by two linear layers with ReLU and dropout.
The results show that introducing image cues unexpectedly degrades overall performance: AMOTA decreases, and FP and IDS increase, although Recall and TP improve and FN decreases.
This observation reflects a recall–robustness trade-off. Image cues increase association flexibility and reduce missed detections, thereby improving Recall. However, under insufficient illumination and severe occlusion (Fig.~\ref{fig:camera_cases}), appearance features become unreliable and introduce noise into the representation space. This noise increases false associations and identity switches, which outweigh the recall gain and reduce AMOTA.
These findings suggest that inference based solely on point cloud and geometric cues is sufficient to achieve robust performance, while avoiding the additional computational overhead and representation interference introduced by image processing.

\subsubsection{Effects of Key Architectural Design}
The positional encoding scheme and the input channels of point cloud data can be configured in multiple ways. Table \ref{tab4} reports the impact of different design choices on the 3D MOT performance evaluated on the nuScenes validation set. 
When positional encoding is implemented using a learnable MLP-based embedding and the input channels consist of $x$, $y$, and $z$, RegTrack achieves an AMOTA of 71.1\%, Recall of 73.8\%, TP of 81907, FP of 13575, FN of 19591, and IDS of 399. When the positional encoding uses a non-learnable sinusoidal formulation, AMOTA, Recall, and TP increase by 1.7\%, 0.9\%, and 537, respectively, while FP, FN, and IDS decrease by 2603, 520, and 17, respectively.
These results indicate that for 3D MOT scenarios with large variations in scene scale, a deterministic sinusoidal positional encoding can preserve precise spatial information and reliably generalize to unseen spatial locations, thereby effectively mitigating the overfitting risk commonly introduced by learnable embeddings.
Furthermore, when the input channels are reduced to $x$ and $y$, a small performance improvement is observed, with AMOTA increasing by 0.2\%. This suggests that in autonomous driving scenarios where tracked objects remain on the ground, the $z$-dimension not only fails to enhance tracking performance but may also introduce redundant information, potentially leading to negative effects.
In summary, when sinusoidal positional encoding is applied to input point cloud patches that contain only the $x$ and $y$ channels, our method achieves the best performance.

In addition, Table \ref{tab5} further analyzes the impact of the number of experts in MoE-GEnc on the 3D MOT performance evaluated on the nuScenes validation set. Compared with using a single expert, employing two experts increases AMOTA, Recall, and TP by 0.5\%, 0.3\%, and 23, respectively, while FP, FN, and IDS decrease by 75, 12, and 11, respectively.
When the number of experts is increased to three, AMOTA decreases by 0.2\%, Recall increases by 0.3\%, and IDS rises by 10. 
Considering model efficiency and the fact that AMOTA is the primary performance metric, a two-expert configuration is selected for MoE-GEnc.

\subsubsection{Effects of LG-PEnc}
LG-PEnc forms point cloud representations by aggregating local and global features and by incorporating hybrid attention to capture fine-grained details.
Table~\ref{tab6} reports the contributions of local features, global features, and hybrid attention to 3D MOT performance on the nuScenes validation set. The results show that the use of local features alone already yields strong performance, achieving an AMOTA of 73.1\%. The inclusion of global features further improves AMOTA by 0.5\%, and the addition of hybrid attention provides an additional gain of 0.2\%.
These results indicate that local features, global features, and hybrid attention each contribute positively to 3D MOT performance.

Table~\ref{tab6-2} further analyzes the impact of different attention normalization strategies for local features (point-wise vs. channel-wise, see Eq.~\ref{eq:local features}) and different aggregation schemes for local and global features (fixed average vs. adaptive weighted, see Eq.~\ref{eq:point cloud embedding}) on 3D MOT performance.
The results indicate that channel-wise attention normalization substantially improves both Recall ($+$0.9\%) and TP ($+$210), while reducing FN ($-$231). These findings suggest that the discriminability of point cloud representations primarily resides in the channel dimension and that each point on an object should be treated equally.
In addition, the results show that adaptive channel-wise weighted aggregation significantly enhances tracking performance, increasing AMOTA by 0.8\% and reducing IDS by 66. This gain further indicates that channel contributions are not uniform and that adaptive channel reweighting is therefore necessary.

In addition, Table~\ref{tab7} further compares the performance of the proposed LG-PEnc with other point cloud encoders on the nuScenes validation set.
The results indicate that LG-PEnc significantly outperforms competing methods, achieving an AMOTA that is 1.7\% higher than that of PointMixer~\cite{pointmixer}. Moreover, LG-PEnc maintains a competitive parameter count compared with other encoders, with the exception of DGCNN\cite{dgcnn}, which introduces fewer convolution operations due to its emphasis on graph construction.
Overall, LG-PEnc achieves strong performance with only 2.6M parameters, demonstrating a favorable trade-off between model complexity and tracking robustness.

\subsection{Efficiency Analysis}
Efficiency is a critical performance metric for 3D MOT. Table \ref{tab8} first analyzes the impact of the three cues employed in UTEnc, namely point cloud, geometric, and image cues, on the parameter count and FLOPs.
The results show that when only the point cloud cue is used, the model contains 2.39 M parameters and requires 0.32 GFLOPs. After incorporating geometric cues, the parameter count increases by only 0.21 M, while the FLOPs remain almost unchanged. In contrast, introducing image cues results in a substantial increase of 86.59 M parameters and an additional 35.17 GFLOPs.
These results indicate that point cloud and geometric cues are highly parameter-efficient and computationally lightweight, requiring far fewer resources than ResNet-18 \cite{resnet} (11.69M parameters, 1.82 GFLOPs for a $3 \times 224 \times 224$ input tensor) and even fewer parameters than PointNet (4.08M).
As a result, excluding image features during inference is highly beneficial for constructing a lightweight 3D MOT model.

Table \ref{tab9} further analyzes the runtime performance of RegTrack on the KITTI and nuScenes test sets, including GPU and CPU memory usage and FPS.
When using only point cloud cues, RegTrack uses 248 MB GPU, 122 MB CPU, and achieves 245 FPS on KITTI, compared to 1011 MB GPU, 138 MB CPU, and 27 FPS on nuScenes. The higher GPU usage and lower FPS on nuScenes are mainly attributable to the greater number of objects per frame.
After integrating geometric cues, GPU and CPU usage increase only slightly, while FPS decreases more noticeably, by 90 FPS on KITTI and 8 FPS on nuScenes. This reduction results from the single-GPU environment, which requires sequential execution of each expert in MoE-GEnc that processes geometric cues. Nevertheless, GPU and CPU usage remain low on both datasets, with inference speed on KITTI substantially exceeding real-time requirements and near real-time performance maintained on nuScenes.
In practical applications, additional acceleration strategies are possible, such as limiting the tracking range to reduce the number of tracked objects or employing parallel or asynchronous computational optimizations.
We further investigate the impact of incorporating an image encoder during inference. On KITTI, GPU memory usage increases substantially, CPU memory remains nearly unchanged, and the frame rate drops significantly. On nuScenes, GPU memory exceeds the available capacity, making inference infeasible.
These results further highlight the efficiency and practical advantage of the proposed method.

Furthermore, RegTrack achieves highly competitive inference speeds across both KITTI and nuScenes datasets. On KITTI, existing multi-modal methods such as mmMOT \cite{mmmot}, JRMOT \cite{jrmot}, MMF‑JDT \cite{mmf-jdt}, and PNAS‑MOT \cite{pnas-mot} operate at only 4 FPS, 14 FPS, 5 FPS, and 13 FPS, respectively, due to their computationally intensive architectures. 

\subsection{Generalizability Analysis}
Existing methods have largely overlooked the generalizability of association metrics and their corresponding thresholds across different object categories and scenarios (i.e., datasets). In contrast, RegTrack adopts a single association metric that is independent of both category and scenario, thereby ensuring strong generalizability.
Table \ref{tab8} analyzes the generalizability of the association threshold on the nuScenes validation set across seven categories.
Varying the threshold from 0.1 to 0.8 on KITTI results in negligible performance changes, with HOTA fluctuating by less than 0.1\%.
On nuScenes, when the threshold ranges from 0.3 to 0.7, performance differences remain small, with AMOTA varying by less than 0.7\%.
Across the seven nuScenes categories, AMOTA exhibits consistent trends: all categories except \textit{Trailer} achieve their best performance at a threshold of 0.5, and for \textit{Trailer}, the performance at 0.5 is only 0.2\% below its optimum.
These results demonstrate that both the association metric and threshold in RegTrack are highly generalizable. A fixed threshold of 0.5 can therefore be applied across different object categories and scenarios without labor-intensive manual tuning.

To further highlight the generalizability of RegTrack, Table~\ref{tab10-2} analyzes the generalizability of the association threshold in the advanced competitor Poly-MOT \cite{poly-mot} on the nuScenes validation set across seven categories.
The results indicate that when a uniform threshold is applied to all seven categories and varied from 1.0 to 1.9, the tracking performance of each category changes substantially, and the optimal performance for each category occurs at a different threshold. As a result, the overall tracking performance consistently falls short of the best performance achieved with class-specific threshold settings.

\begin{figure*}[ht]
    \centering
    \includegraphics[width=\linewidth]{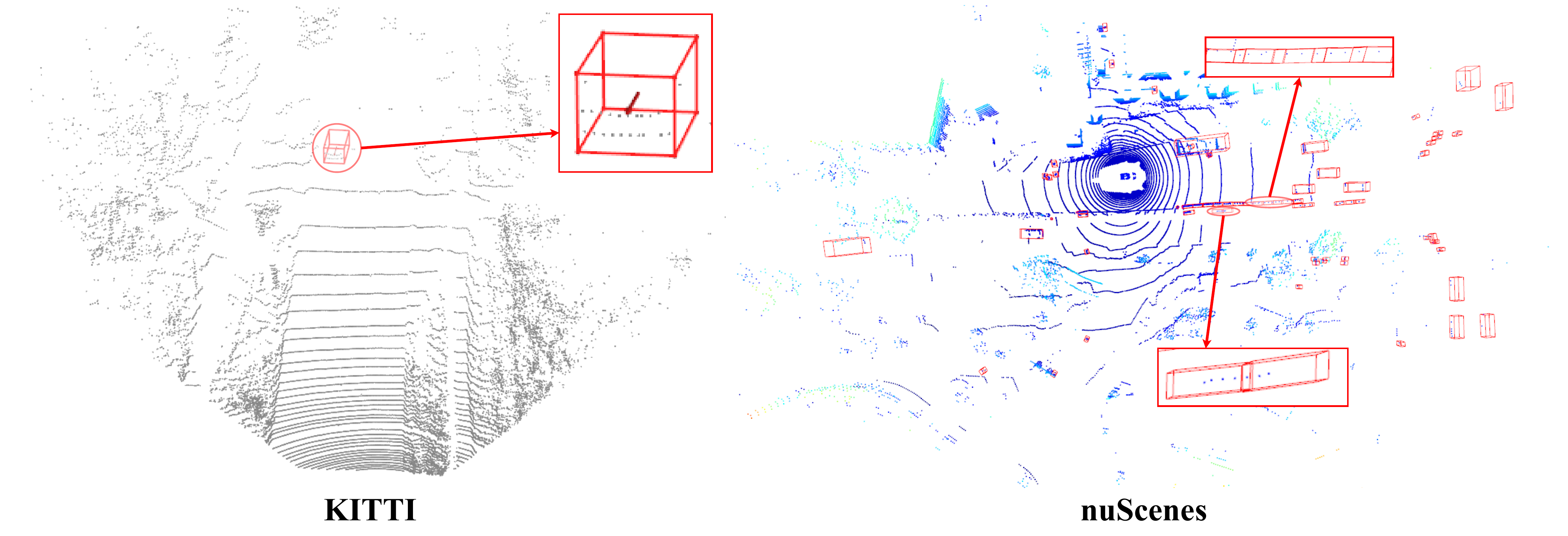}
    \caption{
    \textbf{Visualization of point cloud samples from KITTI and nuScenes.} The left side shows point clouds from the KITTI dataset collected with a 64-beam LiDAR, which are dense, even for distant objects. In contrast, the right side shows point clouds from the nuScenes dataset collected with a 32-beam LiDAR, which are sparse, with only a very limited number of points on slightly distant objects.}
    \label{fig:limitation}
\end{figure*}

In addition, we conduct a cross-domain generalizability analysis, as reported in Table \ref{tab11}. KITTI contains dense point clouds captured with 64 beams, whereas nuScenes provides sparse point clouds captured with 32 beams (Fig. \ref{fig:limitation}).
The results show that training UTEnc on dense point clouds from KITTI and directly testing on sparse point clouds from nuScenes leads to only a slight performance decrease, with an AMOTA drop of 0.6\%. Conversely, training on sparse point clouds from nuScenes and testing on dense point clouds from KITTI yields almost no performance change.
These findings indicate that RegTrack achieves strong cross-dataset and cross-category generalizability using a single association metric with a fixed threshold, and that UTEnc itself exhibits robust generalizability.

\section{Limitations}
Although the above experiments validate the superiority of our RegTrack, it still exhibits certain limitations.
First, while RegTrack exhibits robustness to imperfect 3D detection inputs, its overall performance is still inherently bounded by the limitations of the underlying 3D detectors. As illustrated in Fig. \ref{fig:vis_RegTrack_kitti} (columns 2 to 4), the proposed method does not explicitly address missed detections or provide mechanisms for secondary verification and suppression of false detections. Consequently, detection errors may propagate into the tracking stage, particularly in challenging scenarios with severe occlusion or sensor noise.
Second, since UTEnc primarily relies on point cloud data, its ability to capture discriminative features of distant objects is limited, especially under sparse point cloud conditions. As shown in Fig. \ref{fig:limitation}, in the nuScenes dataset collected with a 32-beam LiDAR sensor, distant objects may be represented by only one or two points, or even no points at all. In such cases, our method currently generates a point at the detection center for objects without point observations. However, this representation is insufficient to encode meaningful appearance or shape cues.
Moreover, RegTrack does not fully exploit temporal appearance evolution across historical frames to infer object appearance changes in future frames. This limitation restricts the ability of RegTrack to maintain robust object representations over long temporal horizons, particularly for distant or intermittently observed targets.

\section{Conclusion}

We introduce RegTrack, the multi-modal 3D object tracking (MOT) framework inspired by Yang–Mills gauge theory. RegTrack redefines the conventional multi-modal paradigm by leveraging an innovative unified tri-cue encoder (UTEnc), achieving strong robustness without sacrificing efficiency or generalizability.
Specifically, UTEnc comprises three encoders: a local–global point cloud encoder (LG-PEnc), a mixture-of-experts-based geometry encoder (MoE-GEnc), and an image encoder derived from a well-pretrained vision–language model (VLM).
LG-PEnc models object point clouds as matter fields, while MoE-GEnc adaptively compensates for inter-frame motions, interpreted as local variations, under the supervision of the composite routing loss. The frozen image encoder is employed exclusively during training, providing a globally invariant reference space, analogous to a physical law, to guide the compensation process under the supervision of the tri-cue unification loss.
Consequently, the motion-compensated point cloud representations of the same object, analogous to observables, maintain global invariance across frames while becoming increasingly discriminative across different objects. RegTrack leverages the pairwise similarities between these representations as an association metric, enabling robust 3D MOT.
Importantly, RegTrack performs inference using only point cloud inputs and does not rely on any class-specific priors, thereby enhancing computational efficiency as well as the generalizability of the association metric and its corresponding threshold.

Our work demonstrates the substantial potential of Yang–Mills gauge theory combined with pretrained VLMs for advancing multi-modal 3D MOT. We anticipate that RegTrack will inspire further research and innovation in the multi-modal 3D MOT domain.


\bibliographystyle{IEEEtran}
\bibliography{acmart}

\vfill

\end{document}